\documentclass[11pt]{article}

\usepackage[preprint]{acl}

\usepackage{times}
\usepackage{latexsym}
\usepackage{mwe}
\usepackage{makecell}  
\usepackage{multirow}
\usepackage{booktabs}
\usepackage[T1]{fontenc}

\usepackage[utf8]{inputenc}
\usepackage{subcaption}
\usepackage{microtype}
\usepackage{amsmath}
\usepackage{amsthm}
\usepackage{amssymb}
\usepackage{tabularx}
\usepackage[table]{xcolor}
\usepackage{titletoc}
\usepackage{inconsolata}
\usepackage[most]{tcolorbox}
\tcbuselibrary{skins, breakable}
\newcolumntype{C}{>{\centering\arraybackslash}X}
\definecolor{mygray}{gray}{0.92}
\usepackage{graphicx}
\definecolor{headergray}{RGB}{85, 85, 85}       
\definecolor{mainbg}{RGB}{245, 245, 245}        
\definecolor{grpoBlue}{RGB}{0, 0, 204}          
\definecolor{vicraMagenta}{RGB}{180, 30, 60}    
\definecolor{wrongRed}{RGB}{204, 50, 50}        
\definecolor{correctGreen}{RGB}{50, 150, 50}    
\definecolor{sourceBox}{RGB}{240, 240, 240}     
\title{Not All Tokens See Equally: Perception-Grounded Policy Optimization for Large Vision-Language Models}

\author{Zekai Ye$^{1}$, Qiming Li$^{1}$, Xiaocheng Feng$^{1,2}$, Ruihan Chen$^1$, \\
\textbf{Ziming Li$^3$, Haoyu Ren$^3$, Kun Chen$^3$, Dandan Tu$^3$, Bing Qin$^{1,2}$}\\
  $^{1}$Harbin Institute of Technology\quad \quad $^2$Peng Cheng Laboratory\quad \quad $^3$Huawei Technologies Co., Ltd\\
  \texttt{\{zkye,qmli\}@ir.hit.edu.cn}
}

\begin{document}
\maketitle
\begin{abstract}
While Reinforcement Learning from Verifiable Rewards (RLVR) has advanced reasoning in Large Vision-Language Models (LVLMs), prevailing frameworks suffer from a foundational methodological flaw: 
by distributing identical advantages across all generated tokens, these methods inherently dilute the learning signals essential for optimizing the critical, visually-grounded steps of multimodal reasoning. 
To bridge this gap, we formulate \textit{Token Visual Dependency}, quantifying the causal information gain of visual inputs via the Kullback-Leibler (KL) divergence between visual-conditioned and text-only predictive distributions. 
Revealing that this dependency is highly sparse and semantically pivotal, we introduce Perception-Grounded Policy Optimization (PGPO), which is a novel fine-grained credit assignment framework that dynamically reshapes advantages at the token level. 
Through a threshold-gated, mass-conserving mechanism, PGPO actively amplifies learning signals for visually-dependent tokens while suppressing gradient noise from linguistic priors. 
Extensive experiments based on the Qwen2.5-VL series across seven challenging multimodal reasoning benchmarks demonstrate that PGPO boosts models by 18.7\% on average. 
Both theoretical and empirical analyses confirm that PGPO effectively reduces gradient variance, prevents training collapse, and acts as a potent regularizer for robust, perception-grounded multimodal reasoning.
Code will be released on https://github.com/Yzk1114/PGPO.
\end{abstract}

\section{Introduction}

\begin{figure}[t]
  \includegraphics[width=\columnwidth]{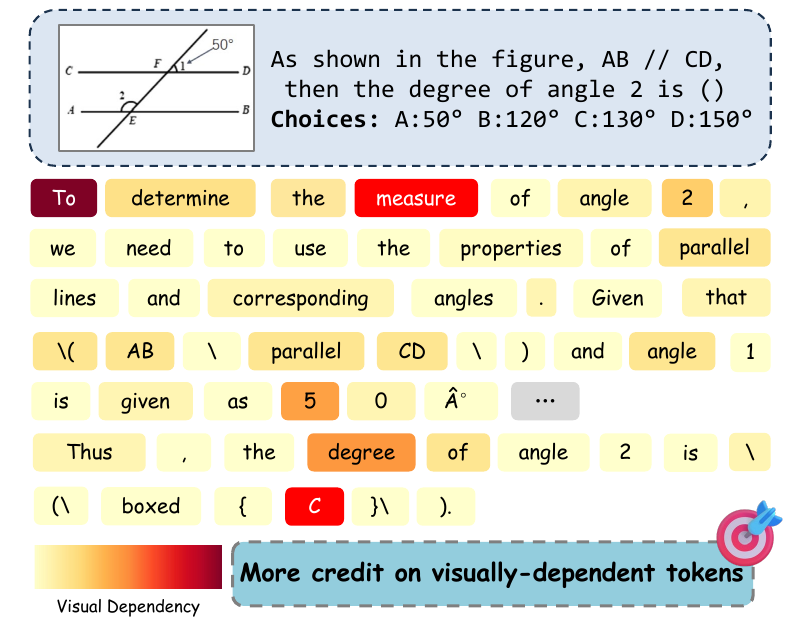}
  \caption{Unlike standard uniform credit assignment, our proposed method dynamically allocates higher reinforcement learning advantage to pivotal tokens that heavily rely on visual perception.}
  \label{fig:intro}
\end{figure}

Reinforcement learning with verifiable rewards (RLVR) \cite{zhang2025survey}—especially online methods such as Group Relative Policy Optimization (GRPO) \cite{shao2024deepseekmath}—has substantially improved reasoning performance in text-oriented Large Language Models (LLMs).
Motivated by these gains, recent studies have extended RLVR to Large Vision-Language Models (LVLMs), with progress mainly driven by improvements in training data design \cite{li2025truth, liang2025modomodo, liu2025noisyrollout, yao2025r1, chen2025g1}, reward construction \cite{shen2025vlm, xia2025visionary, wang2025skywork}, and optimization strategy refinement \cite{wang2025vl, zhao2025absolute,huang2026spotlighttokenperceptionmultimodal}.

Although existing RLVR studies for LVLMs report notable progress, most optimization pipelines still underemphasize fine-grained visual perception.
This creates a key methodological weakness: multimodal reasoning depends on precise perception \cite{xiao2025advancing,wang2025perception}, which serves as the basis for reliable downstream inference. 
In practice, this demand for visual grounding is not uniformly distributed across the generated text.
As illustrated in Figure \ref{fig:intro}, only a critical minority of tokens (e.g., specific geometric entities or numerical values extracted from an image) exhibit high visual dependency. 
Despite this inherent token-level difference, standard GRPO relies on a coarse-grained credit assignment mechanism, uniformly distributing a single sequence-level, group-normalized advantage across all tokens in the generated trajectory.
In the context of multimodal reasoning, this poses a foundational flaw:
by treating all tokens equally, it fundamentally dilutes the learning signals essential for optimizing those pivotal, perception-grounded reasoning steps.

To address this limitation, it is imperative to quantify the causal contribution of the visual modality to each individual reasoning step. 
We formalize this concept as \textit{Token Visual Dependency}, measured by the Kullback-Leibler (KL) divergence between the model's predictive distributions with and without visual conditioning. 
This metric captures the Bayesian surprise \cite{itti2009bayesian}—the specific information gained from the visual context. 
In Section \ref{sec:Preliminaries}, we find this dependency exhibits a highly sparse distribution, demonstrating that visually-grounded reasoning is driven by a critical minority of tokens. 
Furthermore, it reliably isolates semantic visual grounding by assigning higher values to image-dependent visual anchors and truthful concepts.

To this end, we propose \textbf{Perception-Grounded Policy Optimization (PGPO)}, a novel token-level credit assignment framework that integrates threshold-gated, outcome-preserving advantage reallocation into multimodal RLVR. 
PGPO redistributes advantages based on normalized visual dependency scores: it suppresses the gradient noise from low-dependency tokens and actively boosts the learning signal for perceptually pivotal tokens. 
To ensure stable optimization, PGPO applies a sum-preserving normalization that keeps the overall advantage mass consistent with the original, guaranteeing a zero mean for intra-group advantages.

To validate the effectiveness of PGPO, we conduct extensive experiments across seven challenging multimodal reasoning benchmarks.
Based on Qwen2.5-VL \cite{bai2025qwen2} series models, PGPO achieves state-of-the-art performance comparison with previous methods. 
Crucially, by performing fine-grained, token-level advantage reshaping, PGPO reduces gradient variance, accelerates convergence, and increases visual dependency, thereby acting as a potent regularizer for robust, perception-grounded multimodal reasoning.

Our contributions can be summarized as follows:
\begin{itemize}
\item We formulate token visual dependency to quantify the causal impact of visual information, and empirically validate its sparse distribution and correlation with visual semantics.
\item We propose PGPO, a fine-grained credit assignment framework that utilizes token-level advantage reshaping to amplify learning signals for visually-dependent tokens while suppressing noise.
\item Extensive empirical evaluations and theoretical proofs demonstrate that PGPO achieves state-of-the-art performance on diverse multimodal benchmarks, covering mathematical, geometric, logical, multi-discipline reasoning and general VQA.
\end{itemize}

\begin{figure*}
  \centering
  \includegraphics[width=\textwidth]{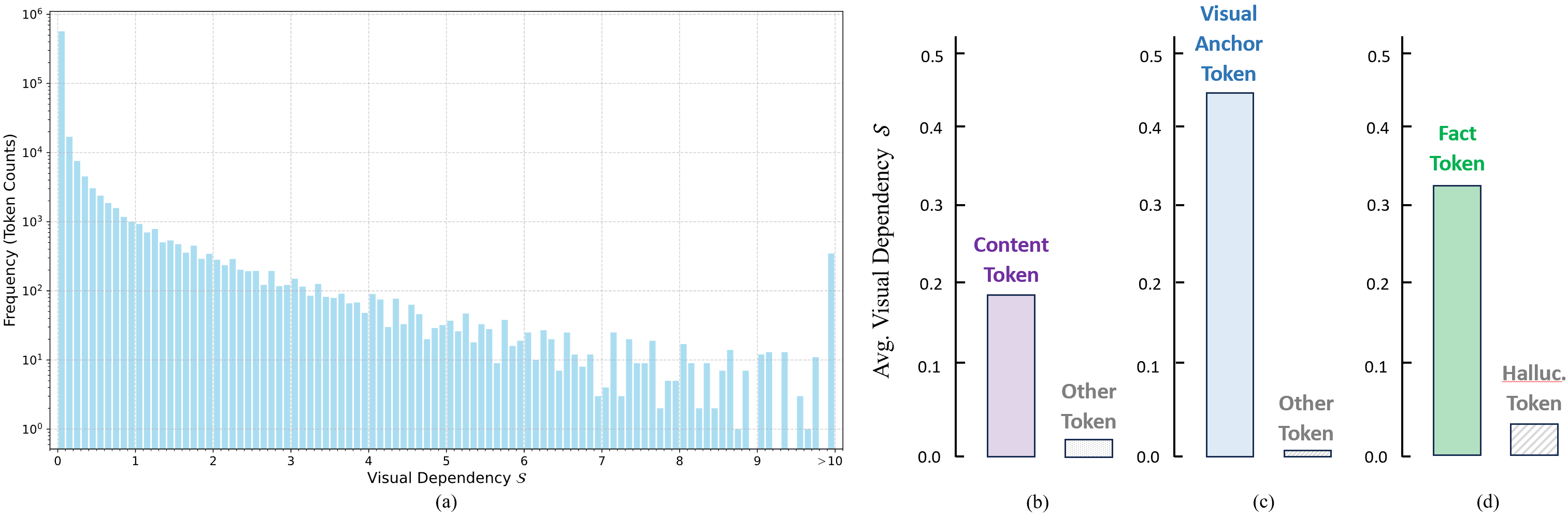}
  \caption{Empirical analysis results of visual dependency. 
  (a) The skewed distribution of token-level visual dependency.
  (b,c,d) Average visual dependency comparison between specific category tokens and others.
  }
  \label{fig:preliminary}
\end{figure*}

\section{Related Work}

\paragraph{Multimodal Reasoning.}
To narrow the remaining performance gap of LVLMs on complex reasoning tasks \cite{wang2024enhancing, dong2025insight}, recent work has extended core reinforcement learning methods such as PPO \cite{schulman2017proximal}, GRPO \cite{shao2024deepseekmath}, and DAPO \cite{yu2025dapo} from text settings to multimodal training \cite{guo2025deepseek, bai2025qwen2, hurst2024gpt}. 
Most existing multimodal RL pipelines, however, improve components around the optimizer rather than the token-level optimization mechanism itself. 
Prior work mainly falls into data-oriented designs \cite{bai2025univg, li2025truth, chen2025advancing, wei2025open, chen2025vinci}
and reward-oriented designs that inject perception-related supervision \cite{wang2025perception, ma2025one}, together with rollout-level changes and external visual tool usage \cite{liu2025noisyrollout, wang2025vl}. 
\citet{huang2026spotlighttokenperceptionmultimodal} improves the optimization of visually dependent tokens through binary gradient filtering, but it lacks dynamic token-level advantage allocation.
Even with these advances, many methods still assign identical training signals to every generated token, which is mismatched with the fine-grained nature of multimodal reasoning.

\paragraph{Credit Assignment.}
Credit assignment \cite{Sutton1998ReinforcementLA, arumugam2021informationtheoreticperspectivecreditassignment, zhou2020learningimplicitcreditassignment} is a central problem in reinforcement learning: determining which earlier actions should receive responsibility for later outcomes. 
This issue becomes more severe in RLVR for LLMs, where sparse terminal rewards must be traced back to token-level decisions across long reasoning trajectories. 
To mitigate this challenge, recent work \cite{wang2025emergenthierarchicalreasoningllms,wei2025reinforcingmultiturnreasoningllm,li2025attentionilluminatesllmreasoning} explores fine-grained attribution for more precise advantage estimation. 
Yet, principled token-level credit redistribution for perception-grounded reasoning in LVLMs remains underexplored.

\section{Preliminaries}
\label{sec:Preliminaries}
To study the perception in multimodal reasoning, this section first formalizes \textit{Token Visual Dependency} via Kullback-Leibler (KL) divergence to quantify the causal information gain from visual inputs at each reasoning step. 
We then empirically demonstrate that this visual dependency exhibits a highly sparse distribution, strongly correlating with semantic visual anchors and factual truthfulness. Together, these theoretical and empirical findings expose the fundamental limitations of uniform sequence-level rewards, directly motivating the core design of our Perception-Grounded Policy Optimization (PGPO) approach.
\subsection{Token Visual dependency}
\label{sec:dependency_quantification}

We characterize token-level visual dependency as the incremental information contributed by the image context, inspired by \citet{huang2026spotlighttokenperceptionmultimodal}. We measure it using the Kullback-Leibler (KL) divergence between the policy's next-token distribution with visual conditioning and the corresponding distribution when visual input is removed, thereby capturing the image-induced distributional change. 

Let $I$ denote the image, $q$ the textual query, and $o$ the generated response. For state $s_t=(q, o_{<t})$, we define the visual dependency score at step $t$ as:
\begin{equation}
    \mathcal{S}(s_t, I) := D_{\text{KL}}\left( \pi_\theta(\cdot | s_t, I) \parallel \pi_\theta(\cdot | s_t) \right).
\end{equation}

A larger $\mathcal{S}$ implies that visual evidence is indispensable for predicting the token at step $t$, indicating a strongly perception-grounded reasoning step.
By quantifying Bayesian surprise—the information gained when updating a language prior to a visually-grounded posterior—KL divergence uniquely isolates the causal intervention of the visual modality on each generated token.
Furthermore, its strict non-negativity, mathematically guaranteed by Gibbs' inequality.
These favorable theoretical properties yield a robust credit assignment signal for reinforcement learning.
More theoretical analyses can be found in Appendix~\ref{sec:kl_proof}, which is fundamentally grounded in the principles of the information theory.

\subsection{Empirical Analysis of Visual Dependency}
\label{sec:empirical_validation}

Having established the theoretical formulation of token-level visual dependency $\mathcal{S}$, we proceed to empirically validate its behavior in multimodal reasoning tasks. 
In this section, we present a comprehensive analysis of its distributional sparsity and categorize tokens based on their roles in the reasoning process to compare their average $\mathcal{S}$ on Qwen2.5-VL-3B.
Together, these empirical findings substantiate $\mathcal{S}$ as a interpretable metric for guiding policy optimization in visual reasoning.
Detailed settings are provided in Appendix~\ref{app:motivation_details}.

\paragraph{Distributional Sparsity of Visual Dependency.}
We further examine the empirical pattern of token-level visual dependency through the proposed metric. Specifically, we run inference on the vision-dominant portion of MathVerse~\citep{zhang2024mathverse} and compute $\mathcal{S}$ for each generated token in every trajectory. 
Figure~\ref{fig:preliminary}(a) shows that, under a logarithmic view, token frequency decreases rapidly as $\mathcal{S}$ becomes larger, revealing a strongly long-tailed distribution. This observation indicates that visually grounded reasoning is driven by only a limited subset of tokens.
In addition, Figure~\ref{fig:preliminary}(b) highlights the semantic relevance of these tokens: content-bearing tokens, such as numbers, geometric concepts, verbs, and adjectives, attain noticeably higher mean $\mathcal{S}$ values than functionally less informative tokens.
Therefore, assigning the same optimization signal to every token is suboptimal, because it spreads credit to many steps that contribute little to actual visual reasoning.

\paragraph{Higher Dependency for Visual Anchors.}
Next, we investigate whether $\mathcal{S}$ is associated with critical visual information. 
We utilize generated trajectories from the MathVerse vision-dominant subset and annotate \textit{visual anchor words} by GPT-5—defined as tokens that are impossible to properly infer without observing the image. 
Figure~\ref{fig:preliminary}(c) reveals that these visual anchor tokens exhibit significantly higher average $\mathcal{S}$ compared to non-anchor tokens. 
This confirms that $\mathcal{S}$ effectively captures semantic visual grounding, making it a meaningful metric for credit assignment to enhance visual reasoning in RLVR.

\paragraph{Lower Dependency in Hallucinations.}
Finally, we explore the correlation between $\mathcal{S}$ and model hallucinations using the CHAIR \cite{rohrbach2019objecthallucinationimagecaptioning} benchmark. 
By generating image captions, we extract and categorize tokens into truthful and hallucinated. 
As shown in Figure~\ref{fig:preliminary}(d), truthful tokens demonstrate higher average $\mathcal{S}$ than hallucinated ones. 
This suggests that hallucinated concepts are primarily driven by the model's language prior rather than actual visual input, further validating $\mathcal{S}$ as a possible proxy for factual visual reliance.

\section{Methodology}

\begin{figure*}
  \centering
  \includegraphics[width=\textwidth]{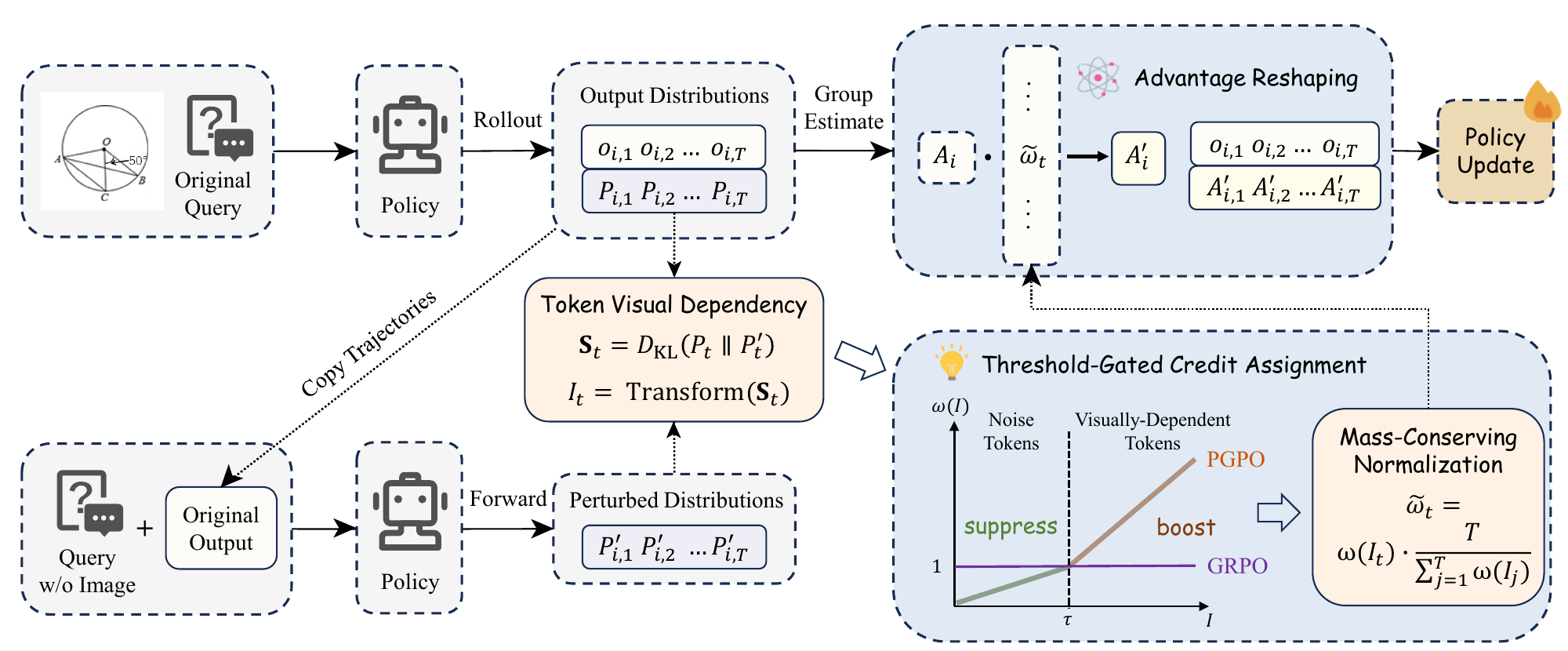}
  \caption{Overview of our proposed PGPO framework. 
  The PGPO pipeline begins by quantifying \textit{token visual dependency} $\mathcal{S}_t$ via KL divergence to isolate the causal information gain from visual inputs. 
  These raw dependency signals are then transformed into bounded \textit{token visual dependency score} $I_t$ through logarithmic compression and min-max normalization. 
  Finally, a threshold-gated mechanism dynamically reshapes the sequence-level GRPO advantage—amplifying learning signals for visually-dependent tokens and suppressing modality-independent noise—while applying a sum-preserving normalization to guarantee stable policy optimization.}
  \label{fig:experiments}
\end{figure*}

Based on above empirical analysis of token visual dependency, to capitalize on fine-grained perceptual signals, we propose Perception-Grounded Policy Optimization (PGPO). It integrates token-level visual dependency with a threshold-gated, mass-conserving advantage reallocation mechanism.

\subsection{Group Relative Policy Optimization}
In the context of multimodal Reinforcement Learning with Verifiable Rewards (RLVR), a Large Vision-Language Model (LVLM) defines an autoregressive policy $\pi_{\theta}$. Given a multimodal prompt $(I, q)$ consisting of a visual input $I$ and a textual query $q$, the old policy $\pi_{\theta_{old}}$ generates a group of $G$ responses, $\{o_{i}\}_{i=1}^{G}$. A reward $R_{i}$ is assigned to each complete response based on whether its final extracted answer matches the ground truth.

GRPO normalizes the outcome reward within the sampled group to compute the advantage $\hat{A}_{i}$ for a response $o_{i}$:

\begin{equation}A_{i}=\frac{R_{i}-mean(\{R_{k}\}_{k=1}^{G})}{std(\{R_{k}\}_{k=1}^{G})}\end{equation}

This single, coarse sequence-level advantage $A_{i}$ is then broadcast indiscriminately to every timestep $t$ in the sequence. The policy $\pi_{\theta}$ is updated by maximizing the clipped surrogate objective:

\begin{align}
& \mathcal{L}_{GRPO}(\theta) = \mathbb{E} \Bigg[ \frac{1}{G} \sum_{i=1}^{G} \frac{1}{|o_{i}|} \sum_{t=1}^{|o_{i}|} \min \Big( \rho_{i,t}(\theta) \\ 
& A_{i}, \notag \text{clip}(\rho_{i,t}(\theta), 1-\epsilon, 1+\epsilon)A_{i} \Big) \Bigg]
\end{align}

where $\rho_{i,t}(\theta)=\frac{\pi_{\theta}(o_{i,t}|I,q,o_{i,<t})}{\pi_{\theta_{old}}(o_{i,t}|I,q,o_{i,<t})}$ represents the probability ratio. However, this uniform assignment neglects the varying contribution of the visual modality of individual steps, resulting in suboptimal visual reasoning and increased gradient variance for tokens that lack visual dependency.

\subsection{Token Visual Dependency Score}

The magnitude of $\mathcal{S}_{t}$ directly quantifies the $t$-th token's reliance on visual grounding.
However, KL divergence is inherently unbounded and typically exhibits a heavy-tailed distribution, making it unsuitable for direct credit assignment. To construct a well-behaved, bounded visual dependency score $I_{t} \in [0, 1]$ for downstream advantage reshaping, we apply a two-stage monotonic transformation.

First, we mitigate the skewed distribution of the raw divergence via a shifted logarithmic compression, defining the dampened signal $\tilde{\mathcal{S}}_{t}$ as:
\begin{equation}\tilde{\mathcal{S}}_{t} = \log(1+\mathcal{S}_{t})\end{equation}
This shift guarantees non-negativity by strictly anchoring zero information gain to zero. More importantly, it prevents extreme outliers from disproportionately dominating the subsequent normalization step, thereby preserving the variance of moderate yet informative attribution signals.

Second, to project these unbounded dampened signals onto a standardized relative scale, we apply sequence-wise min-max normalization:
\begin{equation}I_{t} = \frac{\tilde{\mathcal{S}}_{t} - \min_{i}\tilde{\mathcal{S}}_{i}}{\max_{i}\tilde{\mathcal{S}}_{i} - \min_{i}\tilde{\mathcal{S}}_{i} + \epsilon}\end{equation}

where $i$ indexes the generated sequence, and $\epsilon$ is a small constant preventing division by zero.
This transformation resolves numerical instability and ensures invariance to baseline fluctuations across diverse multimodal prompts.

\subsection{Threshold-Gated Advantage Reshaping}

The execution of a fine multimodal reasoning trajectory involves a interplay between modality-independent logical derivations (yielding low $I_t$) and perception-grounded visual extraction (yielding high $I_t$).
Despite this inherent token-level heterogeneity, standard GRPO uniformly broadcasts a single sequence-level advantage across these diverse tokens, which inevitably dilutes the learning signal for visually grounded reasoning.

Therefore, we propose a threshold-gated advantage reshaping mechanism. 
By introducing the structural prior, we formulate a piece-wise base scaling weight $\omega(I_t)$ that reallocates credit based on each token's visual dependency score:
\begin{equation}
\label{eq:weighting_function}
    \omega(I_t) =
    \begin{cases} 
        \dfrac{I_t}{\tau + \epsilon}, & \text{if } I_t < \tau \\
        1 + \beta \cdot \dfrac{I_t - \tau}{1 - \tau + \epsilon}, & \text{if } I_t \ge \tau
    \end{cases}
\end{equation}

where $\tau \in (0, 1)$ is a relative threshold distinguishing visually-dependent tokens from noise tokens, $\beta \ge 0$ is a boosting coefficient governing the signal amplification regime, and $\epsilon$ is a small constant ensuring numerical stability. 

However, directly modifying the advantage with unbounded scaling weights risks policy update magnitudes, which can destabilize the reinforcement learning optimization. To maintain a consistent update scale, we apply a sum-preserving normalization to the base weights:

\begin{equation}
\tilde{\omega}_{t}=\omega(I_{t})\cdot\frac{|o_{i}|}{\sum_{j=1}^{|o_{i}|}\omega(I_{j})}
\label{eq:sum_preserve}
\end{equation}

where $|o_i|$ denotes the length of the generated response. This normalization mathematically guarantees that the total advantage mass is conserved across the entire trajectory (i.e., $\sum_{t=1}^{|o_i|} \tilde{\omega}_{t} = |o_i|$). 



For the $t$-th token in the $i$-th trajectory, the PGPO advantage is defined as:
\begin{equation}\tilde{A}_{i,t} = A_{i} \cdot \tilde{\omega}_{t}\end{equation}
By substituting $\tilde{A}_{i,t}$ into the GRPO surrogate objective, we derive the final PGPO policy gradient formulation (omitting the clip operation):
\begin{align}
\nabla_\theta \mathcal{L}_{PGPO}(\theta) = \mathbb{E} \Bigg[\rho_{i,t}(\theta) \tilde{A}_{i,t} \nabla_\theta \log \pi_\theta(o_{i,t} | s_{i,t}) \Bigg]
\end{align}


By acting as an active perceptual filter, PGPO amplifies learning signals for visually-dependent tokens while dampening updates for noise tokens. 
This reduces gradient variance and guides the model's capacity toward mastering genuine multimodal reasoning rather than fitting linguistic prior.

\subsection{Theoretical Analysis}
\label{subsec:theoretical_analysis}

In this section, we summarize the theoretical properties of our proposed Perception-Grounded Policy Optimization (PGPO) framework, with detailed proofs provided in the Appendix.

First, our method strictly mitigates the policy gradient variance inherent in GRPO (Appendix \ref{sec:var}). By scaling the gradient contribution of visually-independent tokens by $\varepsilon^2 \ll 1$, it acts as a causal filter that isolates valid reasoning signals and suppresses noise. 

Second, we prove the necessity of advantage mass conservation ($\sum_{t=1}^{|o_i|}\tilde{\omega}_{t}=|o_{i}|$) for training stability (Appendix \ref{sec:sum_preserving_proof}). Unnormalized scaling introduces a non-zero mean shift $\mu$, possibly triggering an $\mathcal{O}(\mu^2)$ explosion in total gradient variance. 

Finally, the mapping from the raw visual dependency score $\mathcal{S}_t$ to the modulated weight $\tilde{\omega}_t$ is strictly monotonically non-decreasing (Appendix \ref{sec:proof_mono}). Furthermore, it preserves the tokens' relative ordinal importance (Appendix \ref{sec:proof_rank}), ensuring credit assignment faithfully reflects visual dependency.

\definecolor{opensourcecolor}{HTML}{fdf3d0}
\definecolor{ourscolor}{HTML}{f2f2fe}
\definecolor{scalingcolor}{HTML}{cadfb8}

\renewcommand{\theadfont}{\scriptsize\bfseries}
\begin{table*}[t]
    \centering
    \footnotesize
    \renewcommand{\arraystretch}{1.25} 
    \setlength{\tabcolsep}{1pt}
        \begin{tabularx}{\textwidth}{C|*{8}{>{\centering\arraybackslash}p{1.4cm}}|>{\centering\arraybackslash}p{1.4cm}} 
            \toprule
            \textbf{Model} & \thead{Geo3k} & \thead{MMK12} & \thead{MathVerse} & \thead{DynaMath} & \thead{MathVision}  & \thead{LogicVista} & \thead{MMMU-Pro} & \thead{$\text{MathVerse}_{V}$} & \textbf{Avg.}\\
            \midrule
            MM-Eureka-7B & 40.30 & 67.68 & 67.15 & 55.01 & 27.13 & 46.30 & 30.30 & 62.41 & 49.54 \\
            VL-Rethinker-7B & 40.70 & 68.30 & 68.80 & 54.97 & 27.91 & 46.47 & 37.13 & 64.96 & 51.16 \\
            NoisyRollout-7B & \textbf{51.98$^*$} & 50.00 & 67.80 & 54.89 & 22.10 & 47.49 & 34.50 & 63.78 & 49.07 \\
            R1-ShareVL-7B & 41.20 & 70.94 & 68.15 & 54.80 & 26.13 & 45.76 & 35.10 & 64.34 & 50.80 \\
            \midrule
            \textit{Qwen2.5-VL-7B} & 37.29 & 43.04 & 38.36 & 48.15 & 17.58 & 43.01 & 26.67 & 33.63 & 35.97 \\
            \qquad\quad\, + GRPO & 30.91 & 76.45 & 68.88 & 54.84 & 27.66 & 46.81 & 36.84 & 64.78 & 50.90 \\
            \qquad\quad\, + DAPO & 41.47 & 78.74 & 67.95 & 55.96 & 28.16 & 47.85 & 36.99 & 64.38 & 52.69 \\
            \qquad\quad\, + PAPO & 43.97 & \underline{80.58} & 69.01 & 56.19 & 28.13 & 47.46 & 38.38 & 64.92 & 53.58 \\
            \qquad\quad\, + VPPO & 44.38 & 80.11 & \underline{70.95} & \underline{57.11} & \underline{28.27} & \underline{47.52} & \underline{38.75} & \underline{65.54} & \underline{54.08} \\
            \rowcolor{ourscolor} \qquad\quad\, + \textbf{PGPO} & \underline{45.20} & \textbf{80.83} & \textbf{71.45} & \textbf{57.71} & \textbf{29.02} & \textbf{47.93} & \textbf{39.01} & \textbf{66.41} & \textbf{54.70} \\
            \midrule
            \midrule
            \textit{Qwen2.5-VL-3B} & 19.65 & 37.37 & 33.16 & 32.88 & 13.88 & 30.17 & 20.95 & 30.31 & 27.30 \\
            \qquad\quad\, + GRPO & 32.19 & 62.95 & 58.39 & 47.82 & 25.59 & 40.85 & 28.02 & 54.36 & 43.77 \\
            \qquad\quad\, + DAPO & 33.80 & 63.36 & 58.76 & 45.86 & 26.19 & 40.63 & 27.20 & 54.89 & 43.84 \\
            \qquad\quad\, + PAPO & 35.11 & 63.54 & 60.81 & 47.30 & 26.37 & \underline{42.70} & 28.76 & 56.90 & 45.19 \\
            \qquad\quad\, + VPPO & \underline{35.36} & \underline{63.76} & \underline{61.34} & \underline{47.75} & \underline{26.40} & 42.03 & \underline{28.83} & \underline{57.02} & \underline{45.31} \\
            \rowcolor{ourscolor} \qquad\quad\, + \textbf{PGPO} & \textbf{36.11} & \textbf{64.74} & \textbf{62.06} & \textbf{48.45} & \textbf{26.90} & \textbf{43.20} & \textbf{29.33} & \textbf{57.24} & \textbf{46.00} \\
            \bottomrule
        \end{tabularx}
    \caption{\textbf{Main Results (avg@8 acc \%).} 
    $\text{MathVerse}_{V}$ refers to the vision-centric subset of MathVerse.
    $^*$NoisyRollout is trained using the training set of Geo3k.}
    \label{tab:main_result}
\end{table*}

\section{Experiments}
\label{sec:experiments}

\subsection{Setups}

\paragraph{Models, Data, and Baselines.}
Consistent with \citet{huang2026spotlighttokenperceptionmultimodal}, we instantiate PGPO on Qwen2.5-VL-3B/7B and train with \texttt{ViRL39K} \cite{wang2025vl}, a heterogeneous dataset of multimodal reasoning tasks. 
For comparison, we evaluate against representative open-source reasoning LVLMs, including MM-Eureka-7B \cite{meng2025mm}, VL-Rethinker-7B \cite{wang2025vl}, R1-ShareVL-7B \cite{yao2025r1}, and NoisyRollout-7B \cite{liu2025noisyrollout}. 
We also reproduce strong multimodal RLVR baselines for GRPO, DAPO, PAPO \cite{wang2025perception}, and VPPO \cite{huang2026spotlighttokenperceptionmultimodal}.

\paragraph{Training Details.}
Following the setup of \citet{huang2026spotlighttokenperceptionmultimodal}, we train for \texttt{2} epochs and a rollout batch size of \texttt{384}. The maximum generation length is set to \texttt{2048}, and the reference KL penalty is removed.
Our PGPO implementation adopts the DAPO training recipe, including dynamic sampling, clip-higher, and token-level policy-gradient optimization.
The bi-level threshold and boosting hyperparameters are fixed to $\tau=0.4$ and $\beta=2.0$, respectively.
Additional implementation details are provided in Appendix~\ref{app:implementation_details}. 

\paragraph{Evaluation Benchmarks.}
We perform a broad evaluation across seven multimodal reasoning benchmarks. 
As in \citet{wang2025perception}, we use exact-match scoring throughout, avoiding dependence on LLM-as-a-judge evaluation. 
These benchmarks cover math, geometry, logic, and multidisciplinary reasoning: DynaMath~\citep{zou2024dynamath}, Geo3k~\citep{lu2021inter}, MathVerse~\citep{zhang2024mathverse}, MathVision~\citep{wang2024measuring}, MMK12~\citep{meng2025mm}, LogicVista~\citep{xiao2024logicvista}, and MMMU-Pro~\citep{yue2024mmmu} (detailed statistics are given in Appendix~\ref{app:benchmark_analysis}). We report average accuracy@8 with inference temperature fixed at 1.0.

\subsection{Main Results}

\paragraph{Strong Performance.}
As detailed in Table~\ref{tab:main_result}, PGPO consistently establishes a new state-of-the-art among existing methods. 
Scaling effectively across both 3B and 7B parameter, our token-level advantage reshaping approach outperforms the next-best baseline, VPPO. 
Crucially, these performance gains are most pronounced on benchmarks demanding rigorous spatial reasoning and high visual dependency, such as Geo3k, MathVerse, and MMMU-Pro. 
While standard sequence-level RL methodologies struggle to assign proper credit during complex, vision-centric problem solving, our method ensures that pivotal visual anchors are accurately rewarded, translating to superior multi-step reasoning capabilities on highly multimodal tasks.

\paragraph{Training Stability.}
\begin{figure}[h]
    \centering
    \begin{subfigure}[b]{0.32\columnwidth}
        \centering
        \includegraphics[width=\columnwidth]{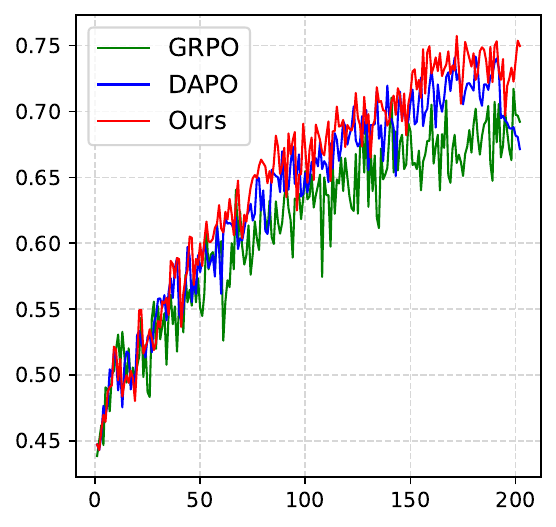}
        \caption{Accuracy.}
        \label{fig:training_accuracy}
    \end{subfigure}
    \hfill 
    \begin{subfigure}[b]{0.30\columnwidth}
        \centering
        \includegraphics[width=\columnwidth]{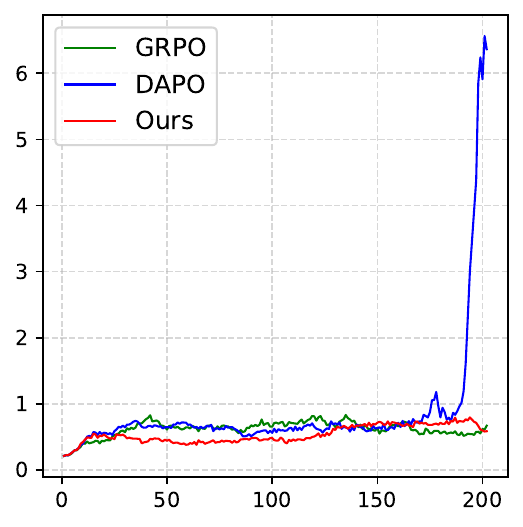}
        \caption{Entorpy.}
        \label{fig:entropy}
    \end{subfigure}
    \hfill
    \begin{subfigure}[b]{0.32\columnwidth}
    \centering
    \includegraphics[width=\columnwidth]{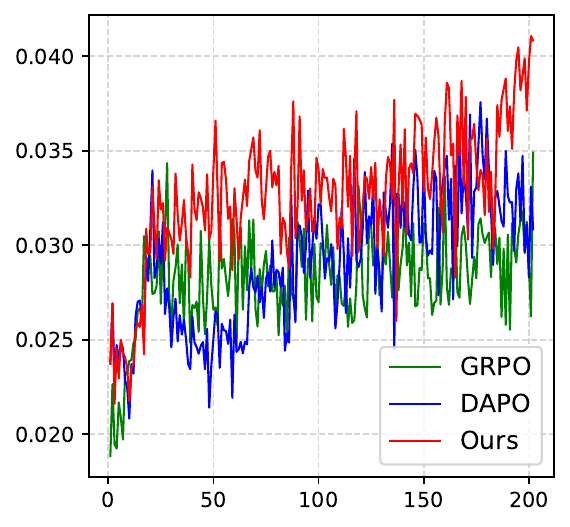}
    \caption{Avg $\tilde{\mathcal{S}}$.}
    \label{fig:vig}
    \end{subfigure}
    \caption{Training dynamics on Qwen2.5-VL-7B.} 
    \label{fig:training_dynamic}
\end{figure}
The effectiveness of PGPO is underpinned by superior training dynamics, as illustrated in the training curves against the baselines (Figure \ref{fig:training_accuracy}), which demonstrates that PGPO exhibits faster initial convergence, achieving higher performance more efficiently.  
Beyond this, another advantage of PGPO lies in its inherent optimization stability. 
As observed in Figure \ref{fig:training_dynamic}, standard approaches like DAPO suffer from severe late-stage training collapse on Qwen2.5-VL-7B, characterized by a sharp drop in accuracy and a corresponding spike in policy entropy (note: DAPO results are reported from the optimal pre-collapse checkpoint to ensure a fair comparison). 
While contemporary baselines such as PAPO and VPPO attempt to mitigate this instability by introducing heuristic entropy penalty terms to their loss functions, PGPO addresses the underlying mathematical root cause. 
By executing fine-grained advantage reallocation for pivot tokens, PGPO reduces gradient variance and stabilizes the training trajectory, eliminating the need for ad-hoc entropy regularization.

\paragraph{Visual Dependency.}
Furthermore, as shown in Figure \ref{fig:vig}, the average visual dependency ($\tilde{\mathcal{S}}_{t}$) of the generated tokens increases alongside the model's training accuracy throughout the RLVR process.
This indicates that the model is fundamentally enhancing its visual perception to facilitate multimodal reasoning.
However, the extent of this perceptual enhancement heavily depends on the underlying optimization strategy.
Under our PGPO framework, visual dependency exhibits a rapid and steady upward trend, whereas baselines like DAPO stagnate during the late training phase. 
This divergence demonstrates that our targeted bi-level learning signal acts as a potent implicit regularizer, actively driving the policy to rely on visual evidence and enabling a highly efficient, robust path to rigorous multimodal reasoning.

\section{Analysis}
\subsection{Effect of the Threshold-Gated Mechanism}
\label{subsec:gating_ablation}

To systematically validate the structural necessity of our proposed threshold-gated advantage reshaping mechanism and its sum-preserving normalization, we conduct an ablation study evaluating isolated variants of PGPO on Qwen2.5-VL-3B.

\begin{table}[ht]
\centering
\resizebox{\linewidth}{!}{
\begin{tabular}{lcc}
\toprule
\textbf{Variant} & \textbf{MathVerse} & \textbf{MMMU-Pro} \\
\midrule
DAPO & 58.76 & 27.20 \\
Suppression-Only ($\tau=1$) & 60.34 & 28.71 \\
Boosting-Only ($\tau=0$)& 59.04 & 28.15 \\
PGPO w/o Norm. & 60.18 & 28.10 \\
\rowcolor{gray!10} \textbf{PGPO} & \textbf{62.06} & \textbf{29.33} \\
\bottomrule
\end{tabular}
}
\caption{Ablation of advantage modulation strategies.}
\label{tab:gating_ablation}
\end{table}

As presented in Table \ref{tab:gating_ablation}, the full PGPO framework substantially outperforms all other variants. 
Deconstructing the gating mechanism reveals that the suppression-only strategy achieves stronger performance than the boosting-only variant. 
This indicates that actively reducing the gradient of visually irrelevant syntactic tokens is more critical for multimodal optimization than merely amplifying perceptual anchors. 

Furthermore, the necessity of Equation \ref{eq:sum_preserve} is empirically verified by the \textit{PGPO w/o Norm.} variant. 
Removing the sum-preserving normalization significantly degrades performance, empirically corroborating our mathematical derivation that unnormalized advantage scaling disrupts policy update constraints via a non-zero mean shift.
Ultimately, integrating threshold-gated suppression, signal boosting, and normalization into the full PGPO configuration yields the optimal synergy.

\subsection{Hyperparameter Analysis}

We further investigate the sensitivity of our method to its core gating hyperparameters on Qwen2.5-VL-3B: the threshold $\tau$ for balancing noise suppression and signal amplification, and the boosting coefficient $\beta$ for scaling score reallocation.

\begin{table}[ht]
\centering
\resizebox{\linewidth}{!}{
\begin{tabular}{cccc}
\toprule
\textbf{Threshold ($\tau$)} & \textbf{Boosting ($\beta$)} & \textbf{MathVerse} & \textbf{MMMU-Pro} \\
\midrule
0.2 & 1.0 & 58.42 & 27.15 \\
0.2 & 2.0 & 59.18 & 27.89 \\
0.4 & 1.0 & 60.36 & 28.98 \\
\rowcolor{gray!10} \textbf{0.4} & \textbf{2.0} & \textbf{62.06} & \textbf{29.33} \\
0.6 & 1.0 & 59.12 & 28.40 \\
0.6 & 2.0 & 56.11 & 28.48 \\
\bottomrule
\end{tabular}
}
\caption{Hyperparameter sensitivity analysis results.}
\label{tab:hyperparameters}
\end{table}

Table \ref{tab:hyperparameters} reveals an optimal configuration at $\tau=0.4$ and $\beta=2.0$. 
A permissive threshold ($\tau=0.2$) dilutes credit assignment by boosting redundant tokens. 
Conversely, a relatively strict threshold ($\tau=0.6$) yields a sparse advantage signal that, when paired with high boosting ($\beta=2.0$), destabilizes PPO clipping bounds and degrades performance. 
Ultimately, $\tau=0.4$ precisely partitions the trajectory, safely isolating the $\beta=2.0$ amplification to true perceptual anchors without inflating global gradients.

\subsection{Performance on General Benchmarks}
\begin{table}[ht]
\centering
\resizebox{\linewidth}{!}{
\begin{tabular}{lccc}
\toprule
\textbf{Method} & \textbf{MMBench} & \textbf{MMStar} & \textbf{MME} \\
\midrule
Qwen2.5-VL-3B & 74.00 & 43.05 & 69.27 \\
GRPO & 83.88 & 57.34 & 75.36\\
DAPO & 84.05 & 57.76 & 75.47 \\
\rowcolor{gray!10} \textbf{PGPO (Ours)} & \textbf{84.96} & \textbf{58.43} & \textbf{77.43} \\
\bottomrule
\end{tabular}
}
\caption{General benchmarks evaluation on Qwen2.5-VL-3B. Performance is reported as avg@8 accuracy.}
\label{tab:general_benchmarks}
\end{table}

To ensure our method does not impair general visual-language capabilities, we evaluated its performance on three unseen, general VQA benchmarks:
MMBench \cite{liu2024mmbenchmultimodalmodelallaround}, MMStar \cite{chen2024rightwayevaluatinglarge} and MME \cite{fu2025mmecomprehensiveevaluationbenchmark}. 

As shown in Table \ref{tab:general_benchmarks}, PGPO consistently outperforms baseline methods, demonstrating a general superiority in vision tasks requiring fine-grained perception or reasoning.
This confirms that anchoring advantage redistribution to token-level visual dependencies prevents the policy from overfitting to superficial textual patterns. 
Meanwhile, PGPO inherently sharpens perceptual grounding without sacrificing general domain knowledge.

\subsection{Analysis of Computational Overhead}
\label{sec:appendix_overhead}

To assess the extra computation introduced by the second forward pass for token visual dependency estimation, we measure training-time overhead relative to the DAPO baseline. 
Table~\ref{tab:overhead_comparison} shows that the added cost is limited to about 10\%. 
This efficiency comes from computing token probabilities for the full sequence in one parallel pass.

In addition, we run DAPO under the same time budget (denoted as $\text{DAPO}_{E}$). 
As reported in Table~\ref{tab:overhead_comparison}, $\text{DAPO}_{E}$ shows little improvement despite longer training, while PGPO delivers a 2-point average gain. 
These results indicate that token-level shaping with visual dependency signals improves learning effectiveness on complex visual reasoning, 
making PGPO's small computational overhead a worthwhile trade-off for performance gains.

\begin{table}[h!]
\centering
\footnotesize
\setlength{\tabcolsep}{3pt}
\renewcommand{\arraystretch}{1.1} 
\renewcommand{\theadfont}{\scriptsize\bfseries}
\resizebox{\linewidth}{!}{
\begin{tabular}{lccc}
\toprule
\thead{Method} & \thead{\begin{tabular}[c]{@{}c@{}}Total Training Time \\ (hours)\end{tabular}} & \thead{Overhead (\%)} & \thead{Avg. Performance} \\ \midrule
{DAPO} & {25.4} & {-} & 43.84 \\
{$\text{DAPO}_{E}$} & \textbf{28.1} & {-} & 43.93 \\
\rowcolor{gray!10} \textbf{{PGPO}} & \textbf{28.1} & \textbf{{+10.6\%}} & \textbf{46.00} \\
\bottomrule
\end{tabular}}
\caption{Analysis results of computational overhead on Qwen2.5-VL-3B with 2 NVIDIA H100 80G GPUs.}
\label{tab:overhead_comparison}
\end{table}

\section{Conclusion}
In this work, we propose Perception-Grounded Policy Optimization (PGPO), which is dynamically reshapes token-level advantages through a threshold-gated, sum-preserving mechanism, actively amplifying learning signals for perceptual anchors while suppressing modality-independent gradient noise. 
Extensive evaluations across seven complex multimodal benchmarks demonstrate the effectiveness of PGPO. 
Both theoretical and empirical analyses confirm that PGPO significantly reduces gradient variance and yields highly robust, perception-grounded multimodal reasoning.

\section*{Limitations}

While PGPO significantly advances multimodal RLVR, we acknowledge certain limitations for future research. 
While the hyperparameters of our threshold-gated mechanism (threshold $\tau$ and boosting coefficient $\beta$) generalize robustly across our benchmarks, they were optimized for our specific training distribution and may require re-tuning when applied to other datasets or model architectures. 
Furthermore, limited by our computational resources, our experiments validate PGPO on models up to the 7B parameter scale. 
Although these results indicate a positive scaling trend, its efficacy on large-scale models remains to be verified. 

\bibliography{custom}
\clearpage
\appendix

\section*{Appendix}
\startcontents[appendix]
\printcontents[appendix]{l}{0}{\setcounter{tocdepth}{1}}

\section{Implementation Details}
\label{app:implementation_details}


\paragraph{Training Details.}
We implement our method on top of the EasyR1 framework \cite{zheng2025easyr1, sheng2024hybridflow}, and run all experiments with PyTorch 2.8.0 and CUDA 12.6. 
Our base models are the open-source Qwen2.5-VL-3B and Qwen2.5-VL-7B checkpoints.
Each model is trained for two epochs on \texttt{ViRL39K} \cite{wang2025vl}, with the vision tower kept trainable throughout optimization. During online RL, we sample 5 responses per prompt. 
The final reward is a weighted sum of accuracy and format compliance, defined as $R = 0.9 S_{\text{acc}} + 0.1 S_{\text{format}}$, where $S_{\text{acc}} \in \{0,1\}$ indicates answer correctness and $S_{\text{format}} \in \{0,1\}$ indicates whether the output satisfies the required response format.
Our objective follows the DAPO training recipe, including dynamic sampling, clip-higher, and token-level policy-gradient optimization, without applying KL regularization. 
Under a unified setup, we also reproduce PAPO and VPPO for direct comparison.
In the main experiments, all runs use the same bi-level threshold $\tau=0.4$ and boosting coefficient $\beta=2.0$ for both training and evaluation.
For fair comparison on Qwen2.5-VL-7B, we report DAPO results from the best pre-collapse checkpoint (step 190 out of 202), since entropy explosion appears near the end of training.
All experiments are conducted on NVIDIA H100 80GB GPU clusters.
Complete optimizer, RL, and evaluation hyperparameters are listed in Table~\ref{tab:hyperparameters}.

\paragraph{Prompt Template.}
Across both training and evaluation, we adopt the standardized prompt template shown below. This format is designed to induce stable Chain-of-Thought reasoning and to support reliable automatic extraction of final answers.

\definecolor{rliableblue}{HTML}{77AADD}
\newcommand{\placeholder}[1]{\textcolor{red}{\{#1\}}}

\begin{tcolorbox}[
    enhanced, 
    title={Reasoning Template},
    fonttitle=\bfseries\sffamily,
    coltitle=white,
    attach boxed title to top center={yshift=-10pt}, 
    boxed title style={
        colback=rliableblue, 
        boxrule=0pt,
        arc=3pt 
    },
    colback=rliableblue!5!white, 
    colframe=rliableblue!80!black, 
    boxrule=0.8pt, 
    arc=4pt, 
    drop shadow=black!5!white, 
    top=12pt, bottom=8pt, left=8pt, right=8pt 
]
\textbf{SYSTEM:} \\ 
You are a helpful assistant. \\[1.5ex] 
\textbf{USER:} \\ 
\placeholder{question} \\ 
You first think through the reasoning process as an internal monologue, enclosed within \texttt{<think> </think>} tags. Then, provide your final answer enclosed within \texttt{\textbackslash boxed\{\}}.
\end{tcolorbox}

\begin{table}[h!]
    \centering
    \caption{Key hyperparameters for training and evaluation.}
    \label{tab:hyperparameters}
    \small
    \begin{tabular}{ll}
        \toprule
        \textbf{Hyperparameter} & \textbf{Value} \\
        \midrule
        \multicolumn{2}{l}{\textit{General Training}} \\
        Optimizer & AdamW \\
        Learning Rate & 1e-6 \\
        LR Schedule & Constant (no warmup or decay) \\
        Epochs & 2 \\
        Freeze Vision Tower & False \\
        \midrule
        \multicolumn{2}{l}{\textit{RL Process}} \\
        Global Batch Size & 128 \\
        Rollout Batch Size & 384 \\
        Rollouts per Prompt & 5 \\
        Rollout Top-p & 0.99 \\
        Max Response Length & 2048 \\
        \midrule
        \multicolumn{2}{l}{\textit{DAPO Recipe}} \\
        Sampling Method & Dynamic Sampling \\
        Clip Ratio Low & 0.2 \\
        Clip Ratio High & 0.28 \\
        Loss Averaging Mode & Token-level \\
        Online Filtering Key & Overall \\
        KL Penalty & None \\
        \midrule
        \multicolumn{2}{l}{\textit{Evaluation Generation}} \\
        Temperature & 1.0 \\
        Top-p & 1.0 \\
        Max New Tokens & 2048  \\
        \bottomrule
    \end{tabular}
\end{table}

\subsection{PGPO Details}

In Section \ref{sec:Preliminaries}, token visual dependency $\mathcal{S}(s_t, I)$ is theoretically defined as the Kullback-Leibler (KL) divergence over the entire vocabulary space $\mathcal{V}$. 
However, computing the exact full-vocabulary KL divergence in Large Vision-Language Models (LVLMs) is practically prohibitive. 
To ensure high training efficiency and stable optimization, we introduce two critical engineering optimizations: attention masking for the unconditioned forward pass, and a low-variance, unbiased Monte-Carlo estimator for the KL divergence.

\paragraph{Visually-Unconditioned Distribution via Attention Masking.}
To obtain the unconditioned probability distribution $\pi_\theta(\cdot | s_t)$, one naive approach is to replace the image with a zero-tensor (blank image) or remove the visual tokens from the input sequence entirely. 
However, altering the input sequence length fundamentally shifts the absolute positional encodings of the subsequent text tokens, introducing confounding variables that distort the pure measurement of visual dependency.

Instead, we compute the unconditioned distribution by manipulating the causal attention mask during a secondary forward pass. Specifically, we set the attention mask corresponding to all visual tokens to \texttt{False}, effectively blinding the text tokens to the visual prefix while preserving their original positional indices. 
It avoids the overhead of re-tokenization and KV-cache structural modifications, allowing the unconditioned logits to be computed seamlessly using the exact same tensor shapes as the primary forward pass.

\paragraph{Estimation of KL Divergence.}
\label{app:k3_estimator}

Mathematically, the exact KL divergence at step $t$ requires computing a sum over the entire vocabulary $\mathcal{V}$ (typically $>100,000$ tokens):
\begin{equation}
    D_{\text{KL}}(\pi_I \parallel \pi_\emptyset) = \sum_{x \in \mathcal{V}} \pi_I(x) \log \frac{\pi_I(x)}{\pi_\emptyset(x)}
\end{equation}
where, for brevity, we denote the visually-conditioned policy as $\pi_I(x) := \pi_\theta(x | s_t, I)$ and the unconditioned policy as $\pi_\emptyset(x) := \pi_\theta(x | s_t)$. 
In practice, explicitly computing this full sum incurs massive GPU memory footprint and I/O overhead. Furthermore, it aggregates the numerical noise from the long tail of highly improbable tokens, which can destabilize the dependency signal.

To resolve this, following \citet{kl_approx}, we approximate the expectation using Monte-Carlo sampling based solely on the generated token $x_t \sim \pi_I$. Let the probability ratio be defined as $r = \frac{\pi_\emptyset(x_t)}{\pi_I(x_t)}$.
We use $(r - 1) - \log r$ as an unbiased, strictly non-negative, and low-variance estimator for KL divergence.

\section{PGPO on the GRPO algorithm}
\definecolor{opensourcecolor}{HTML}{fdf3d0}
\definecolor{ourscolor}{HTML}{f2f2fe}
\definecolor{scalingcolor}{HTML}{cadfb8}

\renewcommand{\theadfont}{\scriptsize\bfseries}
\begin{table*}[ht]
    \centering
    \footnotesize
    \renewcommand{\arraystretch}{1.25} 
    \setlength{\tabcolsep}{1pt}
        \begin{tabularx}{\textwidth}{C|*{8}{>{\centering\arraybackslash}p{1.4cm}}|>{\centering\arraybackslash}p{1.4cm}} 
            \toprule
            \textbf{Model} & \thead{Geo3k} & \thead{MMK12} & \thead{MathVerse} & \thead{DynaMath} & \thead{MathVision}  & \thead{LogicVista} & \thead{MMMU-Pro} & \thead{$\text{MathVerse}_{V}$} & \textbf{Avg.}\\
            \midrule
            \textit{Qwen2.5-VL-3B} & 19.65 & 37.37 & 33.16 & 32.88 & 13.88 & 30.17 & 20.95 & 30.31 & 27.30 \\
            \qquad\quad\, + GRPO & 32.19 & 62.95 & 58.39 & 47.82 & 25.59 & 40.85 & 28.02 & 54.36 & 43.77 \\
            \qquad\quad\, + DAPO & 33.80 & 63.36 & 58.76 & 45.86 & \textbf{26.19} & 40.63 & 27.20 & 54.89 & 43.84 \\
            \rowcolor{ourscolor}  + \textbf{PGPO w/ GRPO} & \textbf{33.81} & \textbf{63.96} & \textbf{59.62} & \textbf{47.91} & 26.00 & \textbf{41.53} & \textbf{28.29} & \textbf{56.75} & \textbf{44.73} \\
            \bottomrule
        \end{tabularx}
    \caption{Results on PGPO by applying it to GRPO (avg@8 acc \%).}
    \label{tab:grpo_ablation}
\end{table*}
To evaluate the generalizability of PGPO, we integrated it with GRPO. As shown in Table~\ref{tab:grpo_ablation}, PGPO also improves GRPO's accuracy, surpassing DAPO. 
This gain aligns with the  improvement observed when applying PGPO to DAPO (Table~\ref{tab:main_result}), confirming that the performance enhancements are intrinsic to our visually-perceptive optimization strategy rather than artifacts of the underlying base policy gradient algorithm.

\section{Detailed Settings for Empirical Analysis of Visual Dependency}
\label{app:motivation_details}

In Section \ref{sec:empirical_validation}, we presented empirical analyses demonstrating the distributional sparsity of token visual dependency ($\mathcal{S}$), its strong correlation with semantic visual anchors, and its inverse relationship with multimodal hallucinations. 
To ensure full reproducibility and provide deeper insights into our methodology, this section details the experimental setups, data processing pipelines, and evaluation metrics used for these preliminary analyses.
All empirical analyses were conducted using the Qwen2.5-VL-3B model. 
To ensure deterministic and reproducible trajectories, we employed greedy decoding for all text generation tasks.

\subsection{Distributional Sparsity and Content Token Analysis}

\paragraph{Dataset Selection.} We utilized the vision-dominant subset of the MathVerse benchmark. This specific subset comprises problems where the visual information is mathematically indispensable, ensuring that a reliance on language priors alone cannot yield the correct answer. We processed the multi-choice test-mini split.

\begin{figure}[!h]
  \includegraphics[width=\columnwidth]{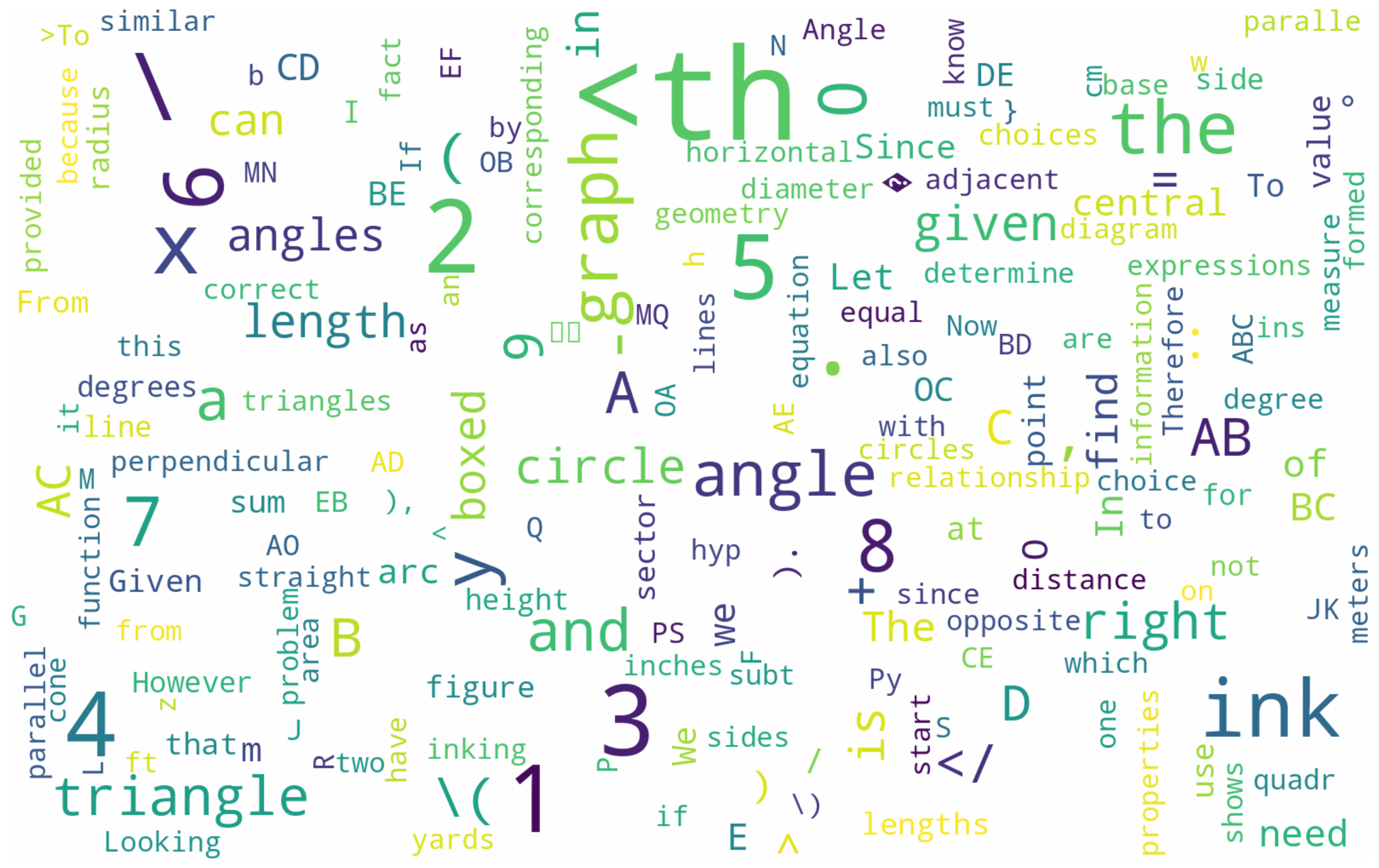}
  \caption{Top 200 $\mathcal{S}$ tokens word cloud on all generated tokens of vision-dominant MathVerse.}
  \label{fig:wordcloud}
\end{figure}

\paragraph{Part-of-Speech (POS) Categorization.} To analyze the discrepancy in visual dependency between different types of tokens, we employed the \texttt{spaCy} to perform POS tagging on the generated trajectories. 
\textit{Content Tokens} were defined as tokens categorized under structural and semantic classes critical to reasoning: `NUM` (numerals), `NOUN` (nouns/geometric concepts), `VERB` (verbs), and `ADJ` (adjectives). \textit{Other Tokens} are other remaining tokens.
Tokens split into sub-words by the tokenizer inherited the POS tag of their parent word. As shown in Figure \ref{fig:preliminary}(b), we computed the mean $\mathcal{S}$ for all tokens within these respective macro-categories across all trajectories.
As shown in Figure \ref{fig:wordcloud}, tokens with high visual dependence often carry critical visual semantics.

\subsection{Visual Anchor Annotation}

To evaluate whether $\mathcal{S}$ is associated with semantic visual grounding, we constructed an annotation pipeline to identify "Visual Anchors"—tokens highly requiring image observation for inference.

\paragraph{Annotation.} We used complete trajectories from the vision-dominant MathVerse generation set. Given the complexity of mathematical reasoning, we leveraged GPT-5 as a zero-shot annotator. GPT-5 was provided with the original image, the textual question, and the generated trajectory, and was prompted to extract specific words that act as visual anchors. 
Prompt for LLM annotation is:
\textit{You are an expert annotator for multimodal mathematical reasoning. Given the following Image, Question, and the Model's Generated Response, your task is to identify 'Visual Anchors' within the response. A Visual Anchor is defined as a specific entity, number, spatial relationship, or geometric attribute mentioned in the text that CANNOT be properly deduced from the text alone and strictly requires extracting information from the image. Output your response as a strictly formatted JSON list of words/phrases extracted directly from the generation.}

\paragraph{Token Mapping.} The extracted anchor phrases were string-matched back to the generated token sequence. Tokens comprising these phrases were labeled as \textit{Visual Anchor tokens}, while all remaining tokens were labeled as \textit{Other tokens}. The statistical significance of the $\mathcal{S}$ differential between these two groups also was confirmed via a two-sample t-test ($p < 0.001$), validating our core hypothesis that high $\mathcal{S}$ corresponds to critical visual grounding.

\subsection{Hallucination Analysis on the CHAIR Benchmark}

To understand the relationship between visual dependency and factual correctness, we extended our analysis to the standard object hallucination benchmark, CHAIR (Caption Hallucination Assessment with Image Relevance) \cite{rohrbach2019objecthallucinationimagecaptioning}.

We sampled 5,000 images from the MSCOCO \cite{lin2015microsoftcococommonobjects} 2014 validation set. For each image, we prompted Qwen2.5-VL-3B with: "Describe this image in detail."
We employed the official CHAIR evaluation script, leveraging MSCOCO ground-truth object annotations (synsets), to assess the generated sequences. 
First, we parsed the generated caption to extract all mentioned objects using the standard CHAIR synonym dictionary. 
Subsequently, these extracted objects are cross-referenced against the image's ground-truth annotations; tokens corresponding to objects present in the ground truth are classified as truthful, whereas those representing absent objects are designated as hallucinated.

We isolated these two subsets of tokens across the 5,000 captions and computed their average $\mathcal{S}$ values. The finding (Figure \ref{fig:preliminary}(d)) that hallucinated tokens exhibit a significantly lower visual dependency provides evidence that multimodal hallucinations in LVLMs are largely driven by over-reliance on language priors rather than visual processing.
This conclusion is consistent with existing research findings \cite{leng2023mitigatingobjecthallucinationslarge,chen2024halcobjecthallucinationreduction,ye2025claimmitigatingmultilingualobject,li2025causaltracingobjectrepresentations,li2025caicaptionsensitiveattentionintervention,li2025unlockingmultilingualreasoningcapability,yu2025rlaifvopensourceaifeedback,zhong2024investigatingmitigatingmultimodalhallucination,chen2025mprguibenchmarkingenhancingmultilingual}.

\section{Information-Theoretic Foundation of the Visual Dependency Metric}
\label{sec:kl_proof}

In this appendix, we establish the theoretical soundness of using the Kullback-Leibler (KL) divergence to quantify visual dependency, denoted as $\mathcal{S}(s_t, I)$. Specifically, we demonstrate that the proposed formulation in Section \ref{sec:dependency_quantification} is mathematically equivalent to the state-specific conditional information gain provided by the visual input.

\subsection{Conditional Mutual Information in Autoregressive Generation}
In information theory, the reduction in uncertainty about a random variable $X$ given the observation of another random variable $Y$ is measured by Mutual Information $I(X; Y)$. When a prior context $Z$ is already known, the additional information provided by $Y$ is captured by the Conditional Mutual Information:
\begin{equation} \label{eq:cmi}
    I(X; Y \mid Z) = H(X \mid Z) - H(X \mid Y, Z),
\end{equation}
where $H(\cdot)$ denotes Shannon entropy. 

Within the context of a Large Vision-Language Model (LVLM), we define the following random variables:
\begin{itemize}
    \item $O_t \in \mathcal{V}$: The next token to be generated from the vocabulary $\mathcal{V}$.
    \item $S_t = (q, o_{<t})$: The current state, comprising the textual query and past generated tokens.
    \item $\mathcal{I}$: The visual input provided to the model.
\end{itemize}

Under the distribution defined by the model policy $\pi_\theta$, the expected information gain regarding the next token $O_t$ provided by the image $\mathcal{I}$, conditioned on the text context $S_t$, is strictly given by $I_{\pi_\theta}(O_t; \mathcal{I} \mid S_t)$.

\subsection{Derivation of Visual Dependency via KL Divergence}
By expanding the definition of Conditional Mutual Information (Equation~\ref{eq:cmi}) using conditional entropy, we can rewrite $I_{\pi_\theta}(O_t; \mathcal{I} \mid S_t)$ in terms of KL divergence. 

Let the visually-conditioned policy be $\pi_\theta(\cdot \mid s_t, I)$ and the visually-marginalized (text-only) policy be $\pi_\theta(\cdot \mid s_t)$. The conditional mutual information over the joint distribution of states and images can be expressed as:
\begin{align}
    & I_{\pi_\theta}(O_t; \mathcal{I} \mid S_t) = \nonumber \\
    & \mathbb{E}_{s_t, I \sim \pi_\theta(S_t, \mathcal{I})} \left[ \sum_{o_t \in \mathcal{V}} \pi_\theta(o_t \mid s_t, I) \log \frac{\pi_\theta(o_t \mid s_t, I)}{\pi_\theta(o_t \mid s_t)} \right] \nonumber \\
    & = \mathbb{E}_{s_t, I \sim \pi_\theta(S_t, \mathcal{I})} \left[ D_{\text{KL}}\Big( \pi_\theta(\cdot \mid s_t, I) \parallel \pi_\theta(\cdot \mid s_t) \Big) \right]. \label{eq:expectation_kl}
\end{align}

Equation~\ref{eq:expectation_kl} demonstrates that the global information gain—averaged across all possible reasoning states and images—is exactly the expected KL divergence between the visually-conditioned policy and the text-only policy.

During RLVR inference, we evaluate the specific informational value of a \textit{given} image instance $I$ at a \textit{given} reasoning state $s_t$, rather than the expectation over the entire distribution. Removing the outer expectation from Equation~\ref{eq:expectation_kl} yields the state-specific conditional information gain:
\begin{align} \label{eq:final_S}
    \mathcal{S}(s_t, I) &:= I_{\pi_\theta}(O_t; \mathcal{I}=I \mid S_t=s_t) \nonumber\\
    & = D_{\text{KL}}\Big( \pi_\theta(\cdot \mid s_t, I) \parallel \pi_\theta(\cdot \mid s_t) \Big).
\end{align}
This confirms that our visual dependency metric $\mathcal{S}(s_t, I)$ fundamentally represents the exact bits of information contributed causally by the image to the prediction of the $t$-th token.

\subsection{Superiority of KL Divergence}
A seemingly intuitive alternative to quantify visual dependency is to compute the direct reduction in entropy before and after observing the image: $\Delta H = H(\pi_\theta(\cdot \mid s_t)) - H(\pi_\theta(\cdot \mid s_t, I))$. However, we exclusively employ KL divergence due to two fundamental theoretical advantages:

\paragraph{Capturing Distributional Shift.} 
Entropy purely measures the \textit{volume} of uncertainty, ignoring the semantic \textit{direction} of the probability shift. Consider a scenario where the text-only policy is highly confident but factually incorrect (e.g., hallucinating a common object based on language priors), yielding a very low entropy $H(\pi_\theta(\cdot \mid s_t))$. When the image is introduced, the model correctly grounds its prediction on visual evidence, yielding a different token but with similarly high confidence (low $H(\pi_\theta(\cdot \mid s_t, I))$). 
In this case, the entropy difference is near zero ($\Delta H \approx 0$), failing to detect the critical intervention of the visual modality. In contrast, KL divergence measures the cross-entropy penalty; because the original text prior assigned low probability to the visually-grounded truth, the KL divergence will be heavily magnified, perfectly capturing the visual correction.

\paragraph{Strict Non-Negativity for Stable Credit Assignment.}
Depending on the clarity of the image, the visual input might actually introduce ambiguity compared to an overconfident language prior, which makes $\Delta H$ negative. Using a metric that can be negative severely complicates advantage reshaping and introduces instability in RL credit assignment. By Gibbs' inequality, KL divergence mathematically guarantees $\mathcal{S}(s_t, I) \ge 0$. This ensures a strict, well-behaved lower bound where $0$ strictly corresponds to identical distributions (zero visual reliance).

\section{Theoretical Analysis of Gradient Noise Suppression}
\label{sec:var}

In vanilla GRPO, the same trajectory-level scalar advantage is broadcast to every decoding position. This design is statistically inefficient when many positions are weakly grounded in the image: those tokens still receive full-magnitude updates even though their contribution to the visual task objective is limited. 
Here we show that PGPO attenuates this irrelevant gradient component and provably reduces the corresponding noise term, which is adapted from \citet{huang2026spotlighttokenperceptionmultimodal}.

\paragraph{Problem Setup and Assumptions.}
Consider one sampled response $y_{1:T}=(y_1,\ldots,y_T)$ under multimodal context $c=(I,q)$. Define the token score vector
\begin{equation}
\mathbf{u}_t := \nabla_\theta \log \pi_\theta(y_t \mid c, y_{<t}).
\end{equation}
Using the dependency indicator $I_t$, we split indices into two subsets:
\begin{equation}
\mathcal{V}:=\{t\,|\, I_t\ge\tau\}, \qquad
\mathcal{B}:=\{t\,|\, I_t<\tau\},
\end{equation}
where $\mathcal{V}$ denotes visually grounded positions and $\mathcal{B}$ denotes background (nuisance) positions.

To characterize estimator noise, we adopt the following assumptions:
\begin{description}
    \item[Assumption 1 (Cross-Time Decorrelation):] For $t\neq j$, $\mathbb{E}[\mathbf{u}_t^\top\mathbf{u}_j]\approx 0$.
    \item[Assumption 2 (Scalar-Gradient Independence):] The trajectory advantage $A$ is independent of $\{\mathbf{u}_t\}_{t=1}^T$.
    \item[Assumption 3 (Weight-Gradient Weak Coupling):] The PGPO weight $\tilde{\omega}_{t}$ is approximately independent of $\mathbf{u}_t$.
    \item[Assumption 4 (Second-Moment Proxy):] In high dimension, $\mathrm{Var}(\mathbf{g})$ is well-approximated by the magnitude of $\mathbb{E}[\|\mathbf{g}\|_2^2]$.
\end{description}

\newtheorem{definition}{Definition}[section]
\newtheorem{theorem}{Theorem}[section]
\newtheorem{proposition}{Proposition}[section]
\newtheorem{lemma}{Lemma}[section]

\begin{theorem}[PGPO Noise Suppression]
\label{thm:PGPO_noise}
Under Assumptions 1--4, if PGPO enforces $\tilde{\omega}_{k}\le \varepsilon$ on $k\in\mathcal{B}$ with $0<\varepsilon\ll1$, then the nuisance-token component in the gradient second moment is reduced by at least a multiplicative factor $\varepsilon^2$ relative to GRPO.
\end{theorem}

\begin{proof}
We compare the squared-norm second moments of the two estimators.

\paragraph{1. GRPO baseline.}
The trajectory gradient in GRPO is
\begin{equation}
\mathbf{g}_{\mathrm{grpo}} = A\sum_{t=1}^{T}\mathbf{u}_t.
\end{equation}
\begin{align}
    & \mathbb{E}\!\left[\|\mathbf{g}_{\mathrm{grpo}}\|_2^2\right]
    = \mathbb{E}\!\left[\left\|A\sum_{t=1}^{T}\mathbf{u}_t\right\|_2^2\right] \nonumber\\
    &\overset{\text{Asm. 2}}{=} \mathbb{E}[A^2]\,\mathbb{E}\!\left[\left\|\sum_{t=1}^{T}\mathbf{u}_t\right\|_2^2\right] \nonumber\\
    &= \mathbb{E}[A^2]\,\mathbb{E}\!\left[\sum_{t=1}^{T}\|\mathbf{u}_t\|_2^2 + \sum_{t\neq j}\mathbf{u}_t^\top\mathbf{u}_j\right] \nonumber\\
    &\overset{\text{Asm. 1}}{\approx} \mathbb{E}[A^2]\sum_{t=1}^{T}\mathbb{E}[\|\mathbf{u}_t\|_2^2] \nonumber\\
    &= \mathbb{E}[A^2]\!\left(\sum_{t\in\mathcal{V}}\mathbb{E}[\|\mathbf{u}_t\|_2^2] + \sum_{k\in\mathcal{B}}\mathbb{E}[\|\mathbf{u}_k\|_2^2]\right). \label{eq:grpo_var_new}
\end{align}
The second summation in Eq.~\ref{eq:grpo_var_new} is pure nuisance contribution: it is nonzero in expectation even when those tokens are weakly related to visual reasoning.

\paragraph{2. PGPO estimator.}
PGPO applies token-wise modulation:
\begin{equation}
\mathbf{g}_{\mathrm{pgpo}} = A\sum_{t=1}^{T}\tilde{\omega}_{t}\mathbf{u}_t.
\end{equation}
Then
\begin{align}
    & \mathbb{E}\!\left[\|\mathbf{g}_{\mathrm{pgpo}}\|_2^2\right]
    = \mathbb{E}\!\left[\left\|A\sum_{t=1}^{T}\tilde{\omega}_{t}\mathbf{u}_t\right\|_2^2\right] \nonumber\\
    &\overset{\text{Asm. 2}}{=} \mathbb{E}[A^2]\,\mathbb{E}\!\left[\sum_{t=1}^{T}\|\tilde{\omega}_{t}\mathbf{u}_t\|_2^2 + \sum_{t\neq j}\tilde{\omega}_{t}\tilde{\omega}_{j}\mathbf{u}_t^\top\mathbf{u}_j\right] \nonumber\\
    &\overset{\text{Asm. 1}}{\approx} \mathbb{E}[A^2]\sum_{t=1}^{T}\mathbb{E}[\tilde{\omega}_{t}^{2}\|\mathbf{u}_t\|_2^2] \nonumber\\
    &\overset{\text{Asm. 3}}{=} \mathbb{E}[A^2]\sum_{t=1}^{T}\mathbb{E}[\tilde{\omega}_{t}^{2}]\,\mathbb{E}[\|\mathbf{u}_t\|_2^2] \nonumber\\
    &= \mathbb{E}[A^2] \nonumber\\
    &\left(\sum_{t\in\mathcal{V}}\mathbb{E}[\tilde{\omega}_{t}^{2}]\mathbb{E}[\|\mathbf{u}_t\|_2^2] + 
    \sum_{k\in\mathcal{B}}\mathbb{E}[\tilde{\omega}_{k}^{2}]\mathbb{E}[\|\mathbf{u}_k\|_2^2]\right). \label{eq:pgpo_var_new}
\end{align}

\paragraph{3. Noise bound and interpretation.} 
By construction, PGPO keeps $\tilde{\omega}_{t}$ around unit scale on $t\in\mathcal{V}$ and enforces $\tilde{\omega}_{k}\le\varepsilon$ on $k\in\mathcal{B}$. Plugging this into Eq.~\ref{eq:pgpo_var_new} gives
\begin{align}
    &\mathbb{E}\!\left[\|\mathbf{g}_{\mathrm{pgpo}}\|_2^2\right]\le \nonumber\\
    &\mathbb{E}[A^2]\left(\sum_{t\in\mathcal{V}}\mathbb{E}[\|\mathbf{u}_t\|_2^2]
    + \varepsilon^2\sum_{k\in\mathcal{B}}\mathbb{E}[\|\mathbf{u}_k\|_2^2]\right).
\end{align}

Combined with Assumption 4, the bound shows a clean separation: the useful visually grounded component is maintained, while the nuisance component is quadratically damped by $\varepsilon^2$ compared with Eq.~\ref{eq:grpo_var_new}. Therefore, in the small-$\varepsilon$ rule, PGPO behaves as a perception-aware variance filter that suppresses updates driven by modality-irrelevant tokens.
\end{proof}

\section{Theoretical Justification for Sum-Preserving Normalization}
\label{sec:sum_preserving_proof}

In the PGPO framework, we dynamically rescale token-level advantages based on visual dependency. However, directly applying unnormalized weights risks fundamentally destabilizing the training process. In this section, we first explain the structural flaw of unnormalized advantage scaling, thereby motivating our sum-preserving normalization mechanism. We then provide a mathematical proof demonstrating the possible variance explosion that occurs if this normalization is omitted.

\subsection{The Flaw of Unnormalized Scaling}
\label{subsubsec:multiplier_limitation}

Let $A_i$ represent the standard GRPO group-normalized sequence advantage for the $i$-th sequence in a sampled group of size $G$. By design, GRPO standardizes rewards such that the advantages maintain a zero mean within the group: $\sum_{i=1}^G A_i = 0$. This implicit zero-sum baseline is critical for reducing the variance of the policy gradient estimator \cite{agarwal2020theorypolicygradientmethods,Kakade2001ANP}.

Suppose we naively introduce unnormalized advantage scaling. Whether applied via a sequence-level coefficient $c_i > 0$ or via token-level weights that do not sum to the sequence length ($\sum_{t=1}^{T_i} \omega_t \neq T_i$), this effectively scales the net advantage of the trajectory, yielding a modulated advantage $\tilde{A}_i = c_i A_i$. Substituting $\tilde{A}_i$ into the standard policy gradient objective, the unclipped gradient estimator becomes:
\begin{align}
    \mathbf{g}_{\text{scaled}} = \frac{1}{G} \sum_{i=1}^{G} \tilde{A}_i \sum_{t=1}^{T_i} \nabla_\theta \log \pi_\theta(o_{i,t} \mid s_{i,t}) \nonumber\\
    = \frac{1}{G} \sum_{i=1}^{G} (c_i A_i) \sum_{t=1}^{T_i} \nabla_\theta \log \pi_\theta(o_{i,t} \mid s_{i,t})
\label{eq:multiplier_grad}
\end{align}

The flaw of this unnormalized scaling lies in the destruction of the zero-mean baseline. With the introduction of the sequence-dependent coefficient $c_i$, the modulated advantages no longer sum to zero:
\begin{equation}
    \mu_{\text{scaled}} = \frac{1}{G} \sum_{i=1}^{G} c_i A_i \neq 0
\end{equation}
This non-zero mean shift ($\mu_{\text{scaled}} \neq 0$) may arbitrarily distorts the magnitude of the parameter updates, disrupting stable credit assignment. 

To prevent this, PGPO employs sum-preserving normalization Equation \ref{eq:sum_preserve}. This mathematically guarantees that the total advantage mass distributed across the trajectory strictly equals $T_i \cdot A_i$, thereby preserving the sequence-level baseline $\mu = 0$ while effectively reallocating credit at the token level.

\subsection{Formal Proof: Asymptotic Quadratic Variance Inflation Induced by Mean Shifts}
\label{subsec:strict_variance_proof}

To rigorously explain why preserving the zero-mean structure is important, we analyze the effect of adding a constant mean shift $\mu$ to an advantage-weighted policy-gradient estimator. The key result is that, although such a shift does not change the estimator's mean, it introduces an additional covariance term that grows quadratically in $|\mu|$ and eventually dominates the estimator variance.

\paragraph{Preliminaries and Definitions.}
Let $(\Omega,\mathcal F,\mathbb P)$ be a probability space, and let $\tau \sim \pi_\theta$ denote a trajectory sampled from a policy $\pi_\theta$ with parameter $\theta \in \mathbb R^d$. Write $\mathbb E[\cdot] := \mathbb E_{\tau \sim \pi_\theta}[\cdot]$.

\begin{description}
    \item[Definition 1 (Score Function).] The trajectory score function is
    $\mathbf g(\tau) := \nabla_\theta \log \pi_\theta(\tau).$
    Under standard regularity conditions,
    $\mathbb E[\mathbf g(\tau)] = \mathbf 0.$

    \item[Definition 2 (Fisher Information Matrix).] The Fisher Information Matrix (FIM) is
    $\mathbf F := \mathbb E\!\left[\mathbf g(\tau)\mathbf g(\tau)^\top\right] \in \mathbb R^{d\times d}.$
    Hence $\mathbf F \succeq 0$. If the policy is non-degenerate in all parameter directions, then $\mathbf F \succ 0$, so $\lambda_{\min}(\mathbf F)>0$.

    \item[Definition 3 (Base and Shifted Estimators).] Let $A^*(\tau)$ be a scalar random variable satisfying
    $\mathbb E[A^*(\tau)] = 0.$
    Define the base estimator
    $\hat{\mathbf g}_0 := A^*(\tau)\mathbf g(\tau),$
    and, for a constant shift $\mu \in \mathbb R$, define the shifted estimator
    $\hat{\mathbf g}_\mu := (A^*(\tau)+\mu)\mathbf g(\tau).$
    
\end{description}
Observe that
\begin{equation}
    \mathbb E[\hat{\mathbf g}_\mu] = \mathbb E[A^*(\tau)\mathbf g(\tau)] + \mu\,\mathbb E[\mathbf g(\tau)] = \mathbb E[\hat{\mathbf g}_0].
\end{equation}
Therefore, adding a constant shift $\mu$ does not change the estimator mean, but it may change its covariance.

\paragraph{Assumption 1 (Finite Moments).}
Assume all expectations below are well-defined and finite. In particular, define
\begin{equation}
\mathbf C := \mathbb E\!\left[A^*(\tau)\mathbf g(\tau)\mathbf g(\tau)^\top\right],
\end{equation}
and assume $\|\mathbf C\|_2 < \infty$ and $\mathrm{tr}(\mathbf C)$ is finite.

\begin{proposition}[Covariance inflation due to a mean shift]
\label{prop:variance_explosion}
Under Assumption 1,
\begin{equation}
\mathrm{Cov}(\hat{\mathbf g}_\mu) = \mathrm{Cov}(\hat{\mathbf g}_0) + \mu^2 \mathbf F + 2\mu \mathbf C.
\end{equation}
Consequently:

\begin{enumerate}
    \item \textbf{Total variance inflation.} If
    \begin{equation}|\mu| > \frac{2|\mathrm{tr}(\mathbf C)|}{\mathrm{tr}(\mathbf F)},\end{equation}
    then
    \begin{equation}\mathrm{tr}\!\left(\mathrm{Cov}(\hat{\mathbf g}_\mu)\right) > \mathrm{tr}\!\left(\mathrm{Cov}(\hat{\mathbf g}_0)\right).\end{equation}

    \item \textbf{Strict inflation in Lowner order.} If $\mathbf F \succ 0$ and
    \begin{equation}|\mu| > \frac{2\|\mathbf C\|_2}{\lambda_{\min}(\mathbf F)},\end{equation}
    then
    \begin{equation}\mathrm{Cov}(\hat{\mathbf g}_\mu) \succ \mathrm{Cov}(\hat{\mathbf g}_0).\end{equation}

    \item \textbf{Asymptotic quadratic dominance.} As $|\mu| \to \infty$,
    \begin{equation}\mathrm{Cov}(\hat{\mathbf g}_\mu)-\mathrm{Cov}(\hat{\mathbf g}_0) = \mu^2\mathbf F + \mathcal O(|\mu|),\end{equation}
    so the covariance penalty is asymptotically dominated by the quadratic term $\mu^2\mathbf F$.
\end{enumerate}
\end{proposition}

\begin{proof}
Since $\mathbb E[\hat{\mathbf g}_\mu]=\mathbb E[\hat{\mathbf g}_0]$, the covariance difference equals the difference of the second-moment matrices:
\begin{align}
\Delta \mathrm{Cov} :&= \mathrm{Cov}(\hat{\mathbf g}_\mu)-\mathrm{Cov}(\hat{\mathbf g}_0) \nonumber\\
&= \mathbb E[\hat{\mathbf g}_\mu \hat{\mathbf g}_\mu^\top] - \mathbb E[\hat{\mathbf g}_0 \hat{\mathbf g}_0^\top].
\end{align}
Expanding $\hat{\mathbf g}_\mu=(A^*+\mu)\mathbf g$, we obtain
\begin{align}
\mathbb E[\hat{\mathbf g}_\mu \hat{\mathbf g}_\mu^\top] =& \mathbb E\!\left[(A^*)^2\mathbf g\mathbf g^\top\right] + 2\mu\,\mathbb E\!\left[A^*\mathbf g\mathbf g^\top\right] + \nonumber\\
&\mu^2\,\mathbb E\!\left[\mathbf g\mathbf g^\top\right].
\end{align}
Recognizing the first term as $\mathbb E[\hat{\mathbf g}_0\hat{\mathbf g}_0^\top]$, the second as $2\mu\mathbf C$, and the third as $\mu^2\mathbf F$, we get
\begin{equation}
    \Delta \mathrm{Cov} = \mu^2\mathbf F + 2\mu\mathbf C.
\end{equation}
For the trace statement,
\begin{equation}\mathrm{tr}(\Delta \mathrm{Cov}) = \mu^2\mathrm{tr}(\mathbf F) + 2\mu\,\mathrm{tr}(\mathbf C).\end{equation}
Using $|2\mu\,\mathrm{tr}(\mathbf C)| \le 2|\mu|\,|\mathrm{tr}(\mathbf C)|$, we have
\begin{equation}\mathrm{tr}(\Delta \mathrm{Cov}) \ge |\mu|\Big(|\mu|\mathrm{tr}(\mathbf F)-2|\mathrm{tr}(\mathbf C)|\Big).\end{equation}
Hence $\mathrm{tr}(\Delta \mathrm{Cov})>0$ whenever
\begin{equation}|\mu| > \frac{2|\mathrm{tr}(\mathbf C)|}{\mathrm{tr}(\mathbf F)}.\end{equation}
For the Lowner-order statement, let $\mathbf v$ be any unit vector. Then
\begin{equation}\mathbf v^\top \Delta \mathrm{Cov}\,\mathbf v = \mu^2 \mathbf v^\top \mathbf F \mathbf v + 2\mu\,\mathbf v^\top \mathbf C \mathbf v.\end{equation}
By Rayleigh-quotient bounds,
\begin{equation}\mathbf v^\top \mathbf F \mathbf v \ge \lambda_{\min}(\mathbf F), \qquad |\mathbf v^\top \mathbf C \mathbf v| \le \|\mathbf C\|_2.\end{equation}
Therefore
\begin{align}
    \mathbf v^\top \Delta \mathrm{Cov}\,\mathbf v &\ge \mu^2\lambda_{\min}(\mathbf F)-2|\mu|\|\mathbf C\|_2 \nonumber\\
    &= |\mu|\Big(|\mu|\lambda_{\min}(\mathbf F)-2\|\mathbf C\|_2\Big).
\end{align}
If
\begin{equation}|\mu| > \frac{2\|\mathbf C\|_2}{\lambda_{\min}(\mathbf F)},\end{equation}
then $\mathbf v^\top \Delta \mathrm{Cov}\,\mathbf v>0$ for every unit vector $\mathbf v$, which implies
\begin{equation}\Delta \mathrm{Cov} \succ 0.\end{equation}
Finally, since
\begin{equation}\Delta \mathrm{Cov} = \mu^2\mathbf F + 2\mu\mathbf C,\end{equation}
the linear term is lower order than the quadratic term as $|\mu|\to\infty$, proving
\begin{equation}\Delta \mathrm{Cov} = \mu^2\mathbf F + \mathcal O(|\mu|).\end{equation}
\end{proof}



\paragraph{Conclusion.}
In very high-dimensional models such as LVLMs, the Fisher term $\mathbf F$ can be large in aggregate, so the quadratic penalty $\mu^2\mathbf F$ may become significant even for modest $|\mu|$. While the exact threshold depends on how both $\mathbf F$ and $\mathbf C$ scale with dimension, the proposition makes clear that non-zero mean shifts become increasingly undesirable in large models because they inject variance in every parameter direction captured by the FIM. This degrades gradient signal-to-noise ratio and can materially destabilize optimization.
This theoretical formulation strictly proves the necessity of maintaining a mathematically exact zero-mean advantage. It serves as the foundational justification for the sum-preserving normalization mechanism proposed in our PGPO framework (Equation \ref{eq:sum_preserve}), ensuring that the modulated advantages fundamentally conserve the $\mu=0$ property while redistributing credit effectively.

\section{Monotonicity Analysis of $\tilde{\omega}_t$}
\label{sec:proof_mono}

\textbf{Proposition G.1 (Monotonicity of Credit Assignment).} \textit{Given a generated sequence, let $\mathcal{S}_{t}$ be the visual dependency score of the $t$-th token. Assuming the scores of all other tokens in the sequence remain constant, the final modulated weight $\tilde{\omega}_t$ is monotonically non-decreasing with respect to $\mathcal{S}_{t}$.}

\begin{proof}
We prove this proposition by systematically analyzing the monotonicity of the four sequential transformations applied to $\mathcal{S}_{t}$: logarithmic compression, min-max normalization, threshold gating, and sum-preserving normalization. 

\paragraph{Step 1: Logarithmic Compression.} 
The first transformation applies a logarithmic dampening: $\tilde{\mathcal{S}}_{t} = \log(1 + \mathcal{S}_{t})$.
Assuming $\mathcal{S}_{t} \ge 0$, the partial derivative is:
\begin{equation}
    \frac{\partial \tilde{\mathcal{S}}_{t}}{\partial \mathcal{S}_{t}} = \frac{1}{1 + \mathcal{S}_{t}} > 0
\end{equation}
Thus, $\tilde{\mathcal{S}}_{t}$ is strictly monotonically increasing with respect to $\mathcal{S}_{t}$.

\paragraph{Step 2: Min-Max Normalization.}
The second transformation standardizes the score:
\begin{equation}
    I_t = \frac{\tilde{\mathcal{S}}_{t} - \min_{j}\tilde{\mathcal{S}}_j}{\max_{j}\tilde{\mathcal{S}}_j - \min_{j}\tilde{\mathcal{S}}_j + \epsilon}
\end{equation}
Because the $\min$ and $\max$ operators evaluate the entire sequence, altering $\tilde{\mathcal{S}}_{t}$ may shift the sequence extrema. Let $m = \min_{j \neq t} \tilde{\mathcal{S}}_j$ and $M = \max_{j \neq t} \tilde{\mathcal{S}}_j$, which act as fixed constants with respect to $\tilde{\mathcal{S}}_{t}$. We analyze $I_t$ across three disjoint intervals:
\begin{enumerate}
    \item \textbf{When $\tilde{\mathcal{S}}_{t} \le m$} (Token $t$ is the minimum): $\min_j \tilde{\mathcal{S}}_j = \tilde{\mathcal{S}}_{t}$. The numerator becomes $\tilde{\mathcal{S}}_{t} - \tilde{\mathcal{S}}_{t} = 0$, yielding $I_t = 0$. The derivative is $0$.
    \item \textbf{When $m < \tilde{\mathcal{S}}_{t} < M$} (Token $t$ is an intermediate value): Here, $\min_j \tilde{\mathcal{S}}_j = m$ and $\max_j \tilde{\mathcal{S}}_j = M$. The derivative is:
    \begin{equation}
        \frac{\partial I_t}{\partial \tilde{\mathcal{S}}_{t}} = \frac{1}{M - m + \epsilon} > 0
    \end{equation}
    \item \textbf{When $\tilde{\mathcal{S}}_{t} \ge M$} (Token $t$ is the maximum): Here, $\max_j \tilde{\mathcal{S}}_j = \tilde{\mathcal{S}}_{t}$ and $\min_j \tilde{\mathcal{S}}_j = m$. Applying the quotient rule:
    \begin{align}
        \frac{\partial I_t}{\partial \tilde{\mathcal{S}}_{t}} &= \frac{\partial}{\partial \tilde{\mathcal{S}}_{t}} \left( \frac{\tilde{\mathcal{S}}_{t} - m}{\tilde{\mathcal{S}}_{t} - m + \epsilon} \right) \nonumber\\
        &= \frac{\epsilon}{(\tilde{\mathcal{S}}_{t} - m + \epsilon)^2} > 0
    \end{align}
\end{enumerate}
Across all intervals, $\frac{\partial I_t}{\partial \tilde{\mathcal{S}}_{t}} \ge 0$. Hence, $I_t$ is monotonically non-decreasing with respect to $\tilde{\mathcal{S}}_{t}$.

\paragraph{Step 3: Threshold-Gating Function.}
The piecewise gating function computes the base weight $\omega_t = \omega(I_t)$. We analyze its piecewise derivatives and boundary continuity:
\begin{enumerate}
    \item \textbf{When $I_t < \tau$}: $\omega_t = \frac{I_t}{\tau + \epsilon} \implies \frac{\partial \omega_t}{\partial I_t} = \frac{1}{\tau + \epsilon} > 0$.
    \item \textbf{When $I_t \ge \tau$}: $\omega_t = 1 + \beta \frac{I_t - \tau}{1 - \tau + \epsilon} \implies \frac{\partial \omega_t}{\partial I_t} = \frac{\beta}{1 - \tau + \epsilon} \ge 0$ (strictly positive for $\beta > 0$).
    \item \textbf{Boundary at $I_t = \tau$}: The right-hand evaluation is $\omega(\tau) = 1$. The left-hand limit is $\lim_{I_t \to \tau^-} \omega(I_t) = \frac{\tau}{\tau + \epsilon}$. Since $\epsilon > 0$, the left limit is strictly less than $1$. This positive jump discontinuity guarantees that the function strictly preserves the non-decreasing property as it crosses the threshold.
\end{enumerate}
Consequently, $\omega_t$ is monotonically non-decreasing with respect to $I_t$.

\paragraph{Step 4: Sum-Preserving Renormalization.}
The final step enforces mass conservation:
\begin{equation}
    \tilde{\omega}_t = \omega_t \cdot \frac{T}{\sum_{j=1}^{T}\omega_j} = \frac{T \cdot \omega_t}{\omega_t + S}
\end{equation}
where $T$ is the sequence length and $S = \sum_{j \neq t} \omega_j > 0$ is a fixed positive constant. Taking the derivative with respect to $\omega_t$:
\begin{equation}
    \frac{\partial \tilde{\omega}_t}{\partial \omega_t} = \frac{T \cdot S}{(\omega_t + S)^2} > 0
\end{equation}
Thus, $\tilde{\omega}_t$ is strictly monotonically increasing with respect to $\omega_t$.
\end{proof}

\paragraph{Conclusion.}
The overall mapping from $\mathcal{S}_t$ to $\tilde{\omega}_t$ is a composition of functions that are differentiable and non-decreasing almost everywhere, with any discontinuities being strictly positive jumps. By the properties of monotonic compositions, the chain strictly preserves order. Therefore, $\tilde{\omega}_t$ is monotonically non-decreasing with respect to $\mathcal{S}_{t}$.
In the context of credit assignment, this mathematical property guarantees that if a specific token demonstrates stronger visual grounding (i.e., higher $\mathcal{S}_{t}$), the PGPO framework will reliably assign it an equal or greater advantage modulation weight $\tilde{\omega}_t$, ensuring stable and interpretable optimization.

\section{Rank-Preserving Property of $\tilde{\omega}_t$}
\label{sec:proof_rank}

\textbf{Proposition H.1 (Intra-Sequence Rank Preservation).} \textit{Given a generated sequence, the PGPO modulation mechanism strictly preserves the relative ordinal importance of the tokens. Specifically, for any two tokens $A$ and $B$ within the same sequence, if their raw visual dependency scores satisfy $\mathcal{S}_A > \mathcal{S}_B$, then their final modulated weights satisfy $\tilde{\omega}_A > \tilde{\omega}_B$.}

\begin{proof}
Let $A$ and $B$ be two distinct tokens in a fixed sequence such that $\mathcal{S}_A > \mathcal{S}_B$. Because both tokens belong to the same sequence, the sequence-level statistics—specifically the unnormalized minimum $m$, maximum $M$, and the pre-normalized sum $S_{\text{total}} = \sum_{j=1}^T \omega_j$—are fixed, shared constants. We evaluate the strict preservation of the inequality through the four transformations:

\paragraph{1. Logarithmic Compression:} Since $f(x) = \log(1+x)$ is strictly monotonically increasing for $x \ge 0$, we have:
\begin{equation}
    \mathcal{S}_A > \mathcal{S}_B \implies \tilde{\mathcal{S}}_A > \tilde{\mathcal{S}}_B
\end{equation}

\paragraph{2. Min-Max Normalization:} The normalization applies a linear transformation:
\begin{equation}
    I_A = \frac{\tilde{\mathcal{S}}_A - m}{M - m + \epsilon}, \quad I_B = \frac{\tilde{\mathcal{S}}_B - m}{M - m + \epsilon}
\end{equation}
Because the denominator $(M - m + \epsilon) > 0$ is identical for both tokens, the order is strictly preserved: $I_A > I_B$.

\paragraph{3. Threshold-Gating Function:} For $\beta > 0$, the piecewise function $\omega(I)$ consists of two linear segments with strictly positive slopes, connected by a positive upward jump at $I = \tau$. Because the function is globally strictly monotonically increasing across its domain $[0, 1]$, we obtain:
\begin{equation}
    I_A > I_B \implies \omega_A > \omega_B
\end{equation}

\paragraph{4. Sum-Preserving Renormalization:} The final step applies sequence-level scaling:
\begin{equation}
    \tilde{\omega}_A = \omega_A \cdot \frac{T}{S_{\text{total}}}, \quad \tilde{\omega}_B = \omega_B \cdot \frac{T}{S_{\text{total}}}
\end{equation}
Since the scalar multiplier $\frac{T}{S_{\text{total}}} > 0$ is a shared constant, the inequality is maintained: $\tilde{\omega}_A > \tilde{\omega}_B$.
\end{proof}

\paragraph{Conclusion.} 
Through all sequential operations, the relationship $\mathcal{S}_A > \mathcal{S}_B \implies \tilde{\omega}_A > \tilde{\omega}_B$ holds strictly true. Therefore, the proposed modulation mechanism mathematically guarantees consistency between a token's empirical visual grounding and its allocated advantage weighting.

\section{Analysis of the Training Dataset}
\label{app:dataset_analysis}

We use \texttt{ViRL39K} \cite{wang2025vl} as the sole training source for reinforcement learning. Its construction matches our objective of learning dependable multimodal reasoning under verifiable supervision.

\paragraph{Breadth of problem domains.}
The dataset contains roughly 39K instances spanning mathematics, physics, chemistry, biology, and chart-centric interpretation. This cross-domain coverage exposes the policy to heterogeneous reasoning patterns and reduces over-specialization to narrow distributions.

\paragraph{Deterministic rewardability.}
Each synthesized sample is paired with a unique, machine-checkable reference answer. This property is central to Reinforcement Learning with Verifiable Rewards (RLVR), where stable optimization depends on objective and automated reward assignment rather than subjective model-as-judge scoring.

\section{Analysis of Evaluation Benchmarks}
\label{app:benchmark_analysis}

Our evaluation protocol uses seven benchmarks to measure complementary abilities: domain-intensive mathematical reasoning, general logic, and cross-discipline multimodal understanding.

\begin{itemize}
    \item \textbf{DynaMath} \cite{zou2024dynamath}: tests robustness by programmatically perturbing numeric values and functional plots in seed questions, helping distinguish memorization from true generalization. We evaluate on \textit{sample variant1}.
    \item \textbf{Geo3k} \cite{lu2021inter}: a large high-school geometry benchmark with formal annotations, suitable for evaluating symbolic and interpretable solution behavior.
    \item \textbf{MathVerse} \cite{zhang2024mathverse}: provides six controlled variants that shift information between text and diagram, enabling fine-grained analysis of visual reliance. We use the test-mini multi-choice subset.
    \item \textbf{MATH-Vision} \cite{wang2024measuring}: sourced from authentic contests (e.g., AMC, Math Kangaroo), spanning 16 subjects and five difficulty levels. We report results on its full test split.
    \item \textbf{MMK12} \cite{meng2025mm}: focuses on K--12 multimodal math and is used to assess foundational reasoning competence.
    \item \textbf{LogicVista} \cite{xiao2024logicvista}: evaluates inductive, deductive, numerical, spatial, and mechanical reasoning across diverse visual formats, offering a broad test of general logical cognition.
    \item \textbf{MMMU-Pro} \cite{yue2024mmmu}: an upgraded MMMU variant designed to suppress text-only shortcuts via stronger option design and vision-centric settings, thereby better testing joint visual-textual reasoning in academic contexts.
\end{itemize}

\section{LLM Usage Statement}
In preparing this manuscript, we used a large language model (LLM) solely to refine language expression, such as correcting grammar and improving sentence fluency. It was not involved in developing the scientific ideas, conducting the analyses, or producing the core technical content. The authors retain full responsibility for all intellectual contributions and for the final version of the paper.

\section{Case Study}
To illustrate how PGPO improves reasoning, we present three representative case studies.
We contrast typical failure patterns of the baseline with the correct reasoning paths produced by PGPO-7B. In all three cases, PGPO-7B answers correctly in all eight sampled generations, demonstrating strong stability and consistency.
These examples highlight the benefit of token-level credit assignment. While a uniform training signal can cause key errors in visual understanding or logical inference, PGPO emphasizes visually grounded critical tokens and guides reasoning toward the correct answer.

\begin{figure*}[htbp]
\begin{tcolorbox}[
    colback=mainbg,          
    colframe=black,          
    boxrule=0.5pt,           
    arc=0mm,                 
    outer arc=0mm,
    sharp corners,
    title=\textcolor{white}{\textbf{\Large Case 1}},
    coltitle=white,
    colbacktitle=headergray,
    fonttitle=\sffamily,
    halign title=left,
    title style={boxrule=0pt}
]

    \begin{tcolorbox}[
        sidebyside,             
        sidebyside align=top,   
        colback=mainbg,         
        boxrule=0pt,            
        lower separated=false,  
        lefthand width=0.42\linewidth 
    ]
        \begin{center}
            \includegraphics[width=0.95\linewidth]{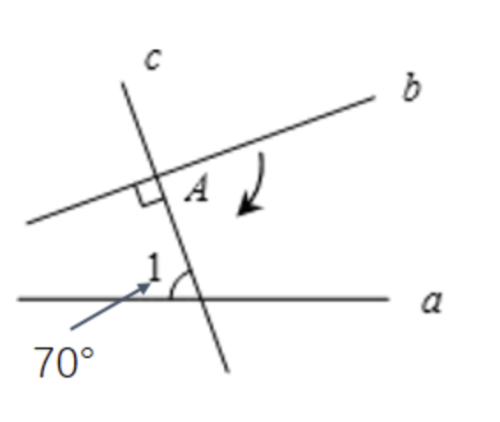} 
        \end{center}
        
        \tcblower 
        
        \begin{minipage}{0.95\linewidth}
            \textbf{\Large Question:} As shown in the figure, If the line b is parallel to the line a, the line b can be rotated () clockwise around the point A 
            
            \vspace{1.5em}
            
            \colorbox{sourceBox}{%
                \parbox{\dimexpr\linewidth-2\fboxsep}{%
                    \texttt{Source: MathVerse, id: 74}%
                }%
            }
            
            \vspace{1.5em}
            
            \textbf{Choices:} (A) 70° \quad (B) 50° \quad (C) 30° \quad (D) 20°

            \vspace{0.5em}
            
            \textcolor{correctGreen}{\textbf{Ground truth:} (D) 20°}
        \end{minipage}
    \end{tcolorbox}

    \vspace{1em} 

    \textcolor{grpoBlue}{\textbf{\sffamily DAPO:}}
    
    \texttt{<think>}
    
    \par\vspace{0.2em}
    Given that \textcolor{red}{line b is parallel to line a}, and an angle measure between them is a given exterior angle at A. Specifically, the given angle adjacent to $\angle$ 1 between line b and the horizontal line is 70°. Since b $\parallel$ a, the angle $\angle$ 1 can be calculated using the supplementary relationship with the adjacent given angle.
So we need to find the rotation angle in order to make b $\parallel$ a. Here the supplementary relation:
$\angle$ 1 + supplementary to it = 180°;
\textcolor{red}{So rotation = 180° - 70°};
    
    \texttt{</think>}
    
    The correct answer is: \fbox{B}
    \quad \textcolor{red}{Wrong}

    \vspace{1.5em}

    \textcolor{vicraMagenta}{\textbf{\sffamily PGPO (Ours):}}
    
    \texttt{<think>}
    
    \par\vspace{0.2em}
   We are required to determine how much line b can rotate clockwise around point A for the lines to maintain the parallel condition.
\textcolor{correctGreen!70!black}{Identify the angles:}
The angle at point A where line b intersects with a horizontal line is given 70°.
Understand the rotation requirement:
We know currently angle 1 is 70°.
For b to parallel to a, rotation is needed.
The current angle effectively dictates that the rotation needed is to move from 70° to a position keeping b parallel to a.
Identify angle at rotation:
Given current alignment requires b shifted from 70° configuration:
Notice the perpendicular creating 90°, angle rotation at A:
Since it currently reads 70°, and b needs to realign, \textcolor{correctGreen!70!black}{rotation = 90° - 70° = 20°}.
Thus, b needs to rotate 20° clockwise around A.
    
    \texttt{</think>}
    
    \vspace{0.5em}
    The correct answer is: \fbox{D} \quad \textcolor{correctGreen}{Correct}

\end{tcolorbox}
\end{figure*}

\begin{figure*}[htbp]
\begin{tcolorbox}[
    colback=mainbg,          
    colframe=black,          
    boxrule=0.5pt,           
    arc=0mm,                 
    outer arc=0mm,
    sharp corners,
    title=\textcolor{white}{\textbf{\Large Case 2}},
    coltitle=white,
    colbacktitle=headergray,
    fonttitle=\sffamily,
    halign title=left,
    title style={boxrule=0pt}
]

    \begin{tcolorbox}[
        sidebyside,             
        sidebyside align=top,   
        colback=mainbg,         
        boxrule=0pt,            
        lower separated=false,  
        lefthand width=0.45\linewidth 
    ]
        \begin{center}
            \includegraphics[width=1\linewidth]{mmmu_case1.png} 
        \end{center}
        
        \tcblower 
        
        \begin{minipage}{1\linewidth}
            \textbf{\Large Question:} (In the image)
            
            \vspace{1.5em}
            
            \colorbox{sourceBox}{%
                \parbox{\dimexpr\linewidth-2\fboxsep}{%
                    \texttt{Source: MMMU-Pro, id: 19}%
                }%
            }
            
            \vspace{1.5em}
            
            \textbf{Choices:} (In the image)

            \vspace{0.5em}
            
            \textcolor{correctGreen}{\textbf{Ground truth:} (E) Skeletal structure}
        \end{minipage}
    \end{tcolorbox}


    \textcolor{grpoBlue}{\textbf{\sffamily DAPO:}}
    
    \texttt{<think>}
    
    \par\vspace{0.2em}
The image shows a floor plan with a grid system, where there are supporting columns or beams at regular intervals. This kind of arrangement suggests a structure where vertical posts (columns) are connected horizontally by lintels (beams).
Given this layout, the architectural system that fits is the \textcolor{red}{Post and lintel system}, which is typically used when there's a need to support a structure with evenly distributed vertical posts and horizontal lintels.
So, among the given options:
(A) Arcuated system - deals with curving structures.
(B) Cantilevering - involves unsupported overhang.
(C) Tensile structure - relies on cables for support.
(D) Post and lintel system - aligns perfectly with evenly distributed supports shown.
(E) Skeletal structure - involves a framework.
(F) Groin vaults - require pointed arches connecting horizontal and vertical systems.
(G) Shell structure - involves curved exterior forms.
(H) Fan vaults - are elaborately used in medieval architecture.
(I) Trabeated system - resembles the post-and-lintel system.

    \texttt{</think>}
    
    The correct answer is: \fbox{D}
    \quad \textcolor{red}{Wrong}

    \vspace{1.5em}

    \textcolor{vicraMagenta}{\textbf{\sffamily PGPO (Ours):}}
    
    \texttt{<think>}
    
    \par\vspace{0.2em}
   The image shows a floor plan with a regular grid of columns and beams. This suggests a structure where supports are aligned in a grid, which is typical for a \textcolor{correctGreen!70!black}{skeletal structure} where the main load-bearing elements are vertical columns and horizontal beams in a grid pattern to support the floor and roof. There is no indication of arches, vaults, or any specific architectural system mentioned in options (A, F, H, I, J). The regular column spacing and arrangement are indicative of a skeletal structure where elements are connected at joints to form a framework that supports the building's structure.
Therefore, the most fitting option here is (E) Skeletal structure.    
    \texttt{</think>}
    
    \vspace{0.5em}
    The correct answer is: \fbox{E} \quad \textcolor{correctGreen}{Correct}

\end{tcolorbox}
\end{figure*}

\begin{figure*}[htbp]
\begin{tcolorbox}[
    colback=mainbg,          
    colframe=black,          
    boxrule=0.5pt,           
    arc=0mm,                 
    outer arc=0mm,
    sharp corners,
    title=\textcolor{white}{\textbf{\Large Case 3}},
    coltitle=white,
    colbacktitle=headergray,
    fonttitle=\sffamily,
    halign title=left,
    title style={boxrule=0pt}
]

    \begin{tcolorbox}[
        sidebyside,             
        sidebyside align=top,   
        colback=mainbg,         
        boxrule=0pt,            
        lower separated=false,  
        lefthand width=0.42\linewidth 
    ]
        \begin{center}
            \includegraphics[width=1\linewidth]{mmmu_case2.png} 
        \end{center}
        
        \tcblower 
        
        \begin{minipage}{1\linewidth}
            \textbf{\Large Question:} (In the image)
            
            \vspace{1.5em}
            
            \colorbox{sourceBox}{%
                \parbox{\dimexpr\linewidth-2\fboxsep}{%
                    \texttt{Source: MMMU-Pro, id: 22}%
                }%
            }
            
            \vspace{1.5em}
            
            \textbf{Choices:} (In the image)

            \vspace{0.5em}
            
            \textcolor{correctGreen}{\textbf{Ground truth:} (H) A book cover}
        \end{minipage}
    \end{tcolorbox}


    \textcolor{grpoBlue}{\textbf{\sffamily DAPO:}}
    
    \texttt{<think>}
    
    \par\vspace{0.2em}
The image shows a beautifully detailed piece that appears to be a flat, decorative work adorned with gemstones and a central figure, suggesting it is meant for display. Given the ornate and decorative style, along with the inclusion of a central iconographic element (the figure of Jesus on the cross), this item likely  \textcolor{red}{served a ceremonial or devotional purpose in a Christian context}.
Reviewing the options:
(A) An icon: Icons are religious artworks primarily used as objects of veneration, matching the devotional and symbolic nature of the artwork.
(B) An altarpiece: Typically a three-dimensional piece, this seems more like a flat display.
(C) A sculpture: Not applicable because the image isn't three-dimensional.
(D), (F), (J): These are also three-dimensional forms, unsuitable for the image.
(E), (G), (H), (I): These forms are not as typical of elaborate decorative flat devotional pieces adorned with jewelry.
Thus, the artwork's nature as a decorative yet religious symbol points toward (A) an icon.
    
    \texttt{</think>}
    
    The correct answer is: \fbox{A}
    \quad \textcolor{red}{Wrong}

    \vspace{1.5em}

    \textcolor{vicraMagenta}{\textbf{\sffamily PGPO (Ours):}}
    
    \texttt{<think>}
    
    \par\vspace{0.2em}
   The image shows a decorative, gem-encrusted piece with a central religious icon of a crucifix, surrounded by ornamental details. This type of artifact is typically used as a cover for a book, especially \textcolor{correctGreen!70!black}{a holy book or a Bible}, given its stylistic and symbolic representation. An icon typically refers to a representation of a sacred subject, often used in Christian religious contexts, which matches the imagery on the piece. The luxurious nature with gems and the central iconography suggest it serves as a protective and valuable cover, often for religious texts like the Bible.
Given the options:
(A) an icon: This usually refers to a small religious image or painting, not necessarily the work shown here.
(B) an altarpiece: Altarpieces are larger and mounted on an altar, which doesn't match the size and style of the object.
(C) a sculpture: The piece is not a three-dimensional sculpture.
(D) a vase design: The piece is not designed to be a vase.
(E) a door ornament: This piece is not a door ornament.
(F) a fresco: Frescoes are painted on walls, which this is not.
(G) a tapestry: While it has ornate detail, it seems too precious to be a tapestry.
(H) a book cover: The ornamented and iconographic style fits a book cover.
(I) a purse lid: This piece is decorative and religious, not a purse lid.
(J) a mural: Murals are typically on larger surfaces like walls, not these kinds of objects.
The most fitting description here is a book cover because of its rich iconographic nature and the way it would be used to cover a book.   
    \texttt{</think>}
    
    \vspace{0.5em}
    The correct answer is: \fbox{H} \quad \textcolor{correctGreen}{Correct}

\end{tcolorbox}
\end{figure*}

\end{document}